\documentclass[10pt]{article}

\usepackage[T1]{fontenc}
\usepackage[utf8]{inputenc}
\usepackage{lmodern}
\usepackage{microtype}
\usepackage{amsmath,amssymb,amsfonts}
\usepackage{graphicx}
\usepackage{booktabs}
\usepackage{threeparttable}
\PassOptionsToPackage{hyphens}{url}  
\usepackage{hyperref}
\hypersetup{
  hidelinks,
  pdftitle={RISED: A Pre-Deployment Evaluation Framework for High-Stakes AI Decision-Support Systems, with Application to Healthcare},
  pdfauthor={Rohith Reddy Bellibatlu, Manpreet Singh, Yash Jajoo, Shyamal Lakhanpal, Abhishek Israni},
  pdfsubject={Pre-Deployment Evaluation of High-Stakes AI Decision-Support Systems},
  pdfkeywords={expert systems, pre-deployment evaluation, trustworthy AI, clinical decision support, algorithmic fairness, model robustness}
}
\usepackage{natbib}
\usepackage{geometry}
\usepackage{setspace}
\usepackage{enumitem}
\usepackage{xcolor}
\usepackage{colortbl}
\usepackage{array}
\usepackage{subcaption}   
\usepackage{etoolbox}     
\usepackage{tikz}
\usetikzlibrary{arrows.meta,positioning}
\usepackage{pgfplots}
\pgfplotsset{compat=1.18}
\usepgfplotslibrary{fillbetween}
\definecolor{cpass}{HTML}{2E7D32}
\definecolor{cfail}{HTML}{C62828}
\definecolor{cincon}{HTML}{E65100}
\definecolor{cbar}{HTML}{1565C0}
\definecolor{cthresh}{HTML}{424242}
\definecolor{ctop3}{HTML}{6A1B9A}

\BeforeBeginEnvironment{tabular}{\setstretch{1.0}}
\BeforeBeginEnvironment{tablenotes}{\setstretch{1.0}}
\BeforeBeginEnvironment{figure}{\setstretch{1.0}}

\definecolor{passbg}{HTML}{CDEBD3}   
\definecolor{failbg}{HTML}{F6D2D0}   
\definecolor{inconbg}{HTML}{FBE6CC}  
\definecolor{diagbg}{HTML}{D4EAF4}   
\newcommand{\verdictFAIL}{\cellcolor{failbg}\textbf{FAIL}}
\newcommand{\verdictPASS}{\cellcolor{passbg}\textbf{PASS}}
\newcommand{\verdictINCONCLUSIVE}{\cellcolor{inconbg}\textbf{INCONCLUSIVE}}
\newcommand{\verdictDIAGNOSTIC}{\cellcolor{diagbg}\textbf{DIAGNOSTIC}}

\title{RISED: A Pre-Deployment Evaluation Framework for\\
       High-Stakes AI Decision-Support Systems,\\
       with Application to Healthcare}

\author{%
  Rohith Reddy Bellibatlu\thanks{%
    Florida International University, Miami, FL, USA.
    \textbf{Corresponding author:} Rohith Reddy Bellibatlu,
    e-mail \texttt{rohithreddybc@gmail.com},
    ORCID \texttt{0009-0003-6083-0364}.}
  \and
  Manpreet Singh\thanks{%
    Boston University, Boston, MA, USA.
    E-mail: \texttt{manni@bu.edu}.
    ORCID: \texttt{0000-0003-2368-2377}.}
  \and
  Yash Jajoo\thanks{%
    New York University, New York, NY, USA.
    E-mail: \texttt{yj1499@nyu.edu}.
    ORCID: \texttt{0009-0005-8103-7415}.}
  \and
  Shyamal Lakhanpal\thanks{%
    University of Maryland, College Park, MD, USA.
    E-mail: \texttt{slakhanp@umd.edu}.
    ORCID: \texttt{0009-0008-3948-511X}.}
  \and
  Abhishek Israni\thanks{%
    Boston University School of Public Health, Boston, MA, USA.
    E-mail: \texttt{abhi1097@bu.edu}.
    ORCID: \texttt{0009-0000-1956-4656}.}
}

\date{}

\begin{document}

\maketitle

\noindent\textbf{Highlights}
\begin{itemize}[leftmargin=2em,topsep=2pt,itemsep=1pt]
  \item Pre-deployment evaluation framework for clinical AI decision-support systems
  \item Five dimensions with bootstrap CIs and PASS/FAIL/INCONCLUSIVE verdicts
  \item Applied to seven clinical cohorts; recurring subgroup and threshold failures
  \item Same failures recur in credit and income prediction, confirming generality
  \item Open-source Python package complementing TRIPOD+AI, FUTURE-AI, and Fairlearn
\end{itemize}

\begin{abstract}
Clinical decision-support systems are expert systems whose recommendations clinicians act on directly, yet they are usually cleared on one aggregate accuracy number from a held-out test set. That number says nothing about input reliability under encoding shifts, subgroup gaps, threshold sensitivity, or operational feasibility. We present \textbf{RISED}, a pre-deployment evaluation framework operationalising five dimensions (\textbf{R}eliability, \textbf{I}nclusivity, \textbf{S}ensitivity, \textbf{E}quity, \textbf{D}eployability) through BCa bootstrap 95\% confidence intervals, literature-grounded thresholds, and Holm--Bonferroni-corrected PASS / FAIL / INCONCLUSIVE verdicts; Equity is a \emph{proxy-dependence diagnostic} rather than a gating test.
Applied to seven cohorts spanning 35 years ($n$ from 303 to 99{,}492), RISED surfaces failures invisible to AUROC: on Diabetes 130, Reliability passes by three orders of magnitude (PSS $=0.0004$) while Inclusivity ($\Delta_\mathrm{AUC}=0.262$) and Sensitivity (max TFR $49.1\%$) fail decisively; both NHIS cohorts reproduce this. NHANES 2021--2023, with a complete feature profile, achieves INCONCLUSIVE verdicts; BRFSS 2024 produces the suite's most severe Sensitivity failure (max TFR $64.2\%$) after instrument rotation removed hypertension and cholesterol. The pattern recurs on credit- and income-prediction cohorts, confirming domain-agnosticity; a multi-model check shows the failures are data-driven, not model-specific.
RISED ships as an open-source Python package complementing TRIPOD+AI, FUTURE-AI, and Fairlearn with the structured numerical evidence those standards require but do not prescribe.
\end{abstract}


\section{Introduction}
\label{sec:introduction}

Expert systems built on machine learning are now embedded in high-stakes clinical decisions across diagnosis, prognosis, triage, and care-management. The dominant pre-deployment evidence is still aggregate discrimination on a held-out test set. That single number is blind to the failure modes that actually matter at deployment: input-encoding instability, silent subgroup degradation, threshold sensitivity, and operational infeasibility. Clinical AI is the domain where the empirical evidence is densest and the regulatory pressure most active, making it the natural test bed for a domain-agnostic evaluation methodology.

Existing toolkits and standards address parts of this problem: AI Fairness 360~\citep{bellamy2019aif360} and Fairlearn~\citep{bird2020fairlearn} package fairness diagnostics; TRIPOD+AI~\citep{collins2024tripodai} and MI-CLAIM~\citep{norgeot2020miclaim} specify what authors must report; CONSORT-AI / SPIRIT-AI~\citep{liu2020consortai,cruzrivera2020spiritai} and FUTURE-AI~\citep{lekadir2025futureai} extend trial reporting and consolidate trustworthy-AI principles. But these resources were built for static model selection or post-hoc reporting, not for determining whether a system is ready to operate reliably and equitably in a live clinical environment~\citep{subbaswamy2021development}.

Deployment introduces conditions that offline benchmarks do not anticipate. Clinically equivalent inputs are often encoded differently across time, clinical site, or EHR system, and the encoding shift alone destabilizes predictions~\citep{finlayson2021clinician}. Underrepresented subpopulations receive systematically degraded predictions while aggregate metrics look clean~\citep{obermeyer2019dissecting,celi2022sources}. Decision thresholds, routinely retuned in deployment to balance sensitivity and specificity, can substantially change which patients the model flags. Clinicians acting on those flags in real time also need interpretable outputs~\citep{sutton2020overview,rudin2019stop}.

These failure modes are documented. The clinical AI literature records systematically degraded subgroup performance under proxy outcomes~\citep{obermeyer2019dissecting}, shortcut learning in imaging diagnostics~\citep{degrave2021ai}, and encoded bias in clinical NLP~\citep{ross2021racial}. The canonical deployment case is the Epic Sepsis Model: externally validated by \citet{wong2021external} on 38{,}455 hospitalizations, it achieved AUROC of only 0.63, missed 67\% of sepsis cases, and generated alerts for 18\% of all hospitalizations despite passing internal benchmarks~\citep{habib2021episfallsshort}. (Epic revised the model in 2022--2023; we cite the 2021 incident as the canonical case of pre-deployment evaluation failing to surface real-world failures.) As \citet{shah2019makingmluseful} argue, accuracy metrics do not measure whether deployment will improve care; pre-deployment evaluation must look beyond aggregate discrimination.

Clinical AI decision-support systems are expert systems in the classical sense: they encode domain knowledge in model parameters and feature pipelines and issue scored recommendations that a domain expert---a clinician---acts upon in a consequential decision. The evaluation challenges they pose are therefore evaluation challenges for expert systems generally.

We propose the \textbf{RISED Framework}, a five-dimension pre-deployment
evaluation approach for clinical AI decision-support systems:
\begin{itemize}[leftmargin=2em]
  \item \textbf{Reliability}: output stability under semantically equivalent but differently encoded inputs;
  \item \textbf{Inclusivity}: performance consistency across demographic subpopulations, via subgroup AUC parity and calibration;
  \item \textbf{Sensitivity}: behavioural stability under decision-threshold shifts, measured via decision flip rates;
  \item \textbf{Equity}: alignment of model predictions with an independent measure of clinical need, beyond demographic parity; and
  \item \textbf{Deployability}: operational feasibility, covering SHAP top-3 feature consistency and end-to-end latency.
\end{itemize}
Each dimension is operationalised through measurable sub-criteria with formal definitions grounded in the published evaluation and fairness literature. The Perturbation Flip Rate and its summary, the Perturbation Sensitivity Score (PSS), adapt input-perturbation robustness measures from clinical machine learning~\citep{finlayson2021clinician,subbaswamy2021development,balendran2025robustness} and adversarial robustness~\citep{madry2018adversarial} to the failure mode of clinically equivalent inputs encoded differently across sites or EHR versions.

\paragraph{Contributions.} This paper makes five independently usable contributions:
\begin{enumerate}[leftmargin=2em,label=\textbf{C\arabic*.}]
  \item \textbf{A domain-agnostic five-dimension evaluation framework}
    for high-stakes AI decision-support systems, organising the
    existing pre-deployment evaluation literature around Reliability,
    Inclusivity, Sensitivity, Equity, and Deployability; we demonstrate
    the domain-agnosticity empirically by reproducing the same failure
    pattern on credit- and income-prediction cohorts
    (\S\ref{sec:application:crossdomain}).
  \item \textbf{A quantitative decision rule} replacing qualitative
    checklists: each dimension is operationalised through formal
    sub-criteria with literature-grounded default thresholds,
    bias-corrected accelerated (BCa) bootstrap 95\% confidence
    intervals, and an explicit PASS / FAIL / INCONCLUSIVE / DIAGNOSTIC
    verdict scheme combined under Holm--Bonferroni family-wise error
    correction.
  \item \textbf{A reframing of Equity as a proxy-dependence diagnostic}
    rather than a stand-alone fairness gate, distinguishing
    statistical demographic parity from need-based alignment in a way
    that resolves a standing tension in the clinical-AI fairness
    literature.
  \item \textbf{An empirical application to seven healthcare cohorts}
    spanning 35 years of data vintage---a 10{,}000-patient synthetic
    cohort, UCI Heart Disease (1989, $n{=}303$), UCI Diabetes 130-US
    Hospitals (1999--2008, $n{=}99{,}492$), NCHS NHIS 2024 Sample
    Adult ($n{=}9{,}747$, cardiovascular), NCHS NHIS 2023 Sample
    Adult ($n{=}27{,}114$, diabetes), NCHS NHANES 2021--2023
    ($n{=}4{,}096$, diabetes), and CDC BRFSS 2024 ($n{=}44{,}888$,
    CHD/MI)---showing that the framework's subgroup and threshold
    failures recur across data settings, outcome types, survey years,
    and feature completeness regimes while its Reliability verdict
    is model-dependent.
  \item \textbf{An open-source Python implementation} (\texttt{rised}),
    released with the cohort generators, evaluation pipelines,
    and a one-command reproducibility kit, so
    that subsequent expert-systems research can re-use the framework
    without reimplementation.
\end{enumerate}

\paragraph{Reusable artefact.} \begin{sloppypar}The implementation is at
\url{https://github.com/rohithreddybc/rised-healthcare-eval} (PyPI release upon acceptance); the synthetic cohort is mirrored on Hugging Face (DOI: \texttt{10.57967/hf/8734}). Both satisfy the FAIR Guiding Principles~\citep{wilkinson2016fair} (persistent DOI, open licence, documented schema, unrestricted reuse) and the TRUST Principles~\citep{lin2020trust} for digital repositories.\end{sloppypar}

\paragraph{Roadmap.} Section~\ref{sec:background} reviews related work and identifies the deployment gap. Section~\ref{sec:framework} specifies all five dimensions. Section~\ref{sec:application} presents clinical-cohort results, multi-model robustness, cross-domain validation on credit and income prediction, and framework coverage. Sections~\ref{sec:discussion}--\ref{sec:conclusion} discuss implications and conclude.

\section{Background and Related Work}
\label{sec:background}

\subsection{AI Evaluation in Healthcare: Tools, Standards, and the Deployment Gap}
\label{sec:background:evaluation}

Clinical machine learning, including generative AI and large language models~\citep{thirunavukarasu2023llm}, has matured into routine deployment across diagnosis, prognosis, and decision support~\citep{rajpurkar2022ai,topol2019highperformance,liu2019reporting}. Several layers of community infrastructure have grown up around these problems. On the toolkit side, AI~Fairness~360~\citep{bellamy2019aif360} and Fairlearn~\citep{bird2020fairlearn,bird2023fairlearn} package fairness diagnostics behind a Python API. Reporting standards fill a different role: TRIPOD+AI~\citep{collins2015tripod,collins2024tripodai}, MI-CLAIM~\citep{norgeot2020miclaim}, CLAIM~\citep{mongan2020claim}, CONSORT-AI / SPIRIT-AI~\citep{liu2020consortai,cruzrivera2020spiritai}, DECIDE-AI~\citep{vasey2022decideai}, and MINIMAR~\citep{hernandezboussard2020minimar} specify what authors must disclose in study reports. Quality-grading instruments then assess those disclosures. PROBAST~\citep{wolff2019probast} is the established risk-of-bias tool for prediction-model studies; APPRAISE-AI~\citep{kwong2023appraiseai} extends similar grading to clinical decision-support AI; FUTURE-AI~\citep{lekadir2025futureai} is an international consensus guideline spanning fairness, universality, traceability, usability, robustness, and explainability. Model cards~\citep{mitchell2019modelcards} and datasheets~\citep{gebru2021datasheets} provide structured-disclosure scaffolding into which RISED's numerical outputs fit. RISED is complementary to all three layers: it adopts the quantitative-gate spirit of PROBAST and APPRAISE-AI but commits to specific metrics, default thresholds, and confidence intervals; it operationalises FUTURE-AI's principles for the subset evaluable pre-deployment; and where DECIDE-AI structures the early live-evaluation report, RISED specifies the numerical pre-deployment evidence on which that evaluation can build. The DOME recommendations~\citep{walsh2021dome} structure supervised ML validation in life sciences across four pillars---Data, Optimization, Model, and Evaluation; RISED extends the Evaluation pillar with deployment-phase threshold tests, BCa bootstrap CIs, and a formal decision rule for high-stakes expert systems beyond the biology domain.

A structural gap remains. The dominant evaluation paradigm was built for static model selection: evaluation happens once, on a held-out test set from the same data-generating process as training. The resulting metrics answer whether the model ranks patients well \emph{within the development sample}, but say nothing about behaviour at a different site, EHR version, or shifted population~\citep{subbaswamy2021development,kelly2019key,ghassemi2020review}. \citet{kelly2019key} document a systematic gap between development-phase accuracy and deployment-phase reliability that existing evaluation practice is not set up to detect. The ML Test Score~\citep{breck2017mltestscore} operationalises production readiness as a rubric of infrastructure, data-pipeline, and model-serving tests; it does not include statistical fairness tests, subgroup calibration, need-based equity, or threshold-sensitivity analysis. The deployment-challenges survey of \citet{paleyes2022challenges} maps organisational and technical barriers ML systems face between research and production; RISED's five dimensions operationalise quantitative detection of the statistical and operational barriers in that taxonomy before a system reaches production.

\subsection{Fairness, Equity, and Bias in Clinical AI}
\label{sec:background:fairness}

Health disparities are amplified by clinical AI systems.
Obermeyer et al.~\citep{obermeyer2019dissecting} showed that a widely
deployed risk score systematically routed Black patients away from
care-management programs because its target variable (healthcare spending)
was depressed for groups with constrained access. Similar training-objective
accuracy masking subgroup harm recurs in radiological deep
learning~\citep{degrave2021ai}, clinical NLP~\citep{ross2021racial}, and
sex-stratified prediction~\citep{cirillo2020sex}, with systematic
reviews~\citep{nazer2023bias,celi2022sources} confirming the pattern is
structural. Population-level outcome heterogeneity across age, sex, and
race/ethnicity~\citep{osibogun2024adverse} reinforces this from a
public-health direction.

The same failure pattern recurs across high-stakes expert-systems domains. Fair-lending audits in credit scoring document systematic disparate impact that aggregate AUC does not surface~\citep{mehrabi2021survey,bartlett2022consumer}. Reviews of algorithmic hiring tools report subgroup selection-rate gaps violating the EEOC four-fifths rule even when overall accuracy looks acceptable~\citep{raghavan2020mitigating}. The ProPublica analysis of COMPAS found false-positive disparities by race that aggregate recalibration metrics failed to flag~\citep{angwin2016machine}. In all three domains, aggregate discrimination is necessary but not sufficient for safe deployment.

Formal fairness criteria (equalized odds~\citep{hardt2016equality},
group-conditional calibration~\citep{pleiss2017fairness}, individual
fairness~\citep{dwork2012fairness}) capture different moral intuitions about
predictive equality; \citet{barocas2023fairness} consolidate the
mathematical landscape in textbook form, and \citet{chen2023algorithmic}
survey how these criteria translate into algorithmic-fairness practice
in medicine and healthcare specifically. A well-known impossibility result~\citep{chouldechova2017fair} shows these criteria are mutually inconsistent when group base rates differ, and \citet{friedler2019comparative} demonstrates empirically that different fairness-intervention algorithms produce substantively different verdicts on the same data---an argument for reporting multiple metrics with CIs rather than collapsing to a single fairness number. \citet{paulus2020predictably} draw the clinically critical distinction: \emph{statistical demographic parity} differs from \emph{need-based fairness}, i.e., predictions that track actual disease burden rather than utilisation proxies distorted by access barriers. \citet{liu2023translational} sharpen this: in clinical deployment, \emph{equity}---not statistical \emph{equality}---is the appropriate target, and metric choices must be clinically motivated. Fairness fixes that work in-distribution often fail to generalize~\citep{yang2024limits}, and penalizing group-fairness violations during training degrades within-group performance~\citep{pfohl2021empiricalfairness}---motivating RISED's Inclusivity dimension reporting $\Delta\mathrm{AUC}$ and ECE alongside the aggregate verdict. \citet{raji2019actionableaudit} demonstrate that publicly naming quantified bias results in commercial AI systems prompted vendor remediation, establishing that auditing produces practical accountability value beyond academic reporting; this motivates RISED's binary PASS/FAIL/INCONCLUSIVE verdict structure over continuous-score approaches that obscure whether a threshold has been met.

\subsection{Reporting Standards and Regulatory Context}
\label{sec:background:standards}

Regulatory pressure is now pushing in the same direction across multiple jurisdictions. In the United States, the FDA AI/ML SaMD Action Plan~\citep{fda2021aiml} introduces predetermined change-control plans for adaptive models, while the ONC HTI-1 rule~\citep{onc2024hti1} mandates that algorithmic decision-support tools expose inputs, logic, and subgroup performance to clinicians. The EU AI Act~\citep{euaiact2024} takes a broader approach, classifying AI used to inform clinical decisions as high-risk and triggering conformity assessment, transparency, and human-oversight obligations across member states. In Germany, the Works Constitution Act (\textit{Betriebsverfassungsgesetz}) grants Betriebsr{\"a}te (works councils) co-determination rights over algorithmic monitoring tools; clinical AI deployments in German hospitals may therefore require works-council agreement alongside EU AI Act conformity assessment, adding a labour-law layer that purely technical evaluation frameworks do not address.

Trial-stage standards address the same disclosure problem prospectively:
CONSORT-AI~\citep{liu2020consortai} and SPIRIT-AI~\citep{cruzrivera2020spiritai}
extend CONSORT/SPIRIT to trials of AI interventions;
DECIDE-AI~\citep{vasey2022decideai} structures early-stage live evaluation
upstream of those trials; MINIMAR~\citep{hernandezboussard2020minimar}
defines minimum model information (training population, target population,
architecture, validation procedure) and is conceptually closest to the
disclosure scope RISED's per-dimension report fills with numerical content.
These frameworks specify what must appear in a study report; none specifies what numerical bar a candidate system must clear \emph{before} such a study is warranted---that pre-study gate is what RISED targets, a gap named in governance scholarship~\citep{reddy2020governance} and the AMIA consensus statement on AI-enabled clinical decision support~\citep{labkoff2024cdss}. The organisational dimension is addressed by \citet{raji2020accountability}, who define an end-to-end internal auditing process (scoping, artefact collection, testing, reflection, review) and explicitly identify the absence of standardised quantitative test batteries as the primary operationalisation gap; RISED provides that battery, producing bootstrap-CI-backed verdicts that feed into such an audit without replacing its governance process. At the policy level, the NIST AI Risk Management Framework~\citep{nist2023aiirmf} structures AI risk across four functions---Govern, Map, Measure, Manage---and calls for quantitative evaluation evidence in its Measure function without prescribing specific metrics or thresholds; RISED targets precisely that gap.

\subsection{Gaps in Existing Frameworks}
\label{sec:background:gaps}

These bodies of work leave four gaps that RISED targets directly; each gap motivates one dimension in Section~\ref{sec:framework}.

\textbf{First}, one-shot held-out evaluation says nothing about how a model behaves when the same clinical reality is encoded differently: a diagnosis at a coarser ICD granularity, a lab result in different units, or a comorbidity flag from a slightly different SQL query can all change the input vector for a patient whose clinical state is unchanged~\citep{finlayson2021clinician,subbaswamy2021development,zhang2022shifting,wong2021external}. The case for testing this is well established~\citep{finlayson2021clinician,subbaswamy2021development,balendran2025robustness}; TEHAI~\citep{reddy2021tehai_bmj} argues deployment readiness deserves its own phase. External validation, while necessary, is not sufficient for deployment readiness: \citet{kovacheva2025external} show that a high-performing postpartum haemorrhage model fell to AUROC 0.60 when externally validated on 87{,}662 deliveries, and required local refitting---which recovered discrimination to 0.75---before clinical use. Missing from the toolkit layer is a packaged, threshold-bearing encoding-stability test that integrates cleanly with the rest of a clinical AI evaluation report.

\textbf{Second}, most fairness audits stop at aggregate demographic parity without asking whether predictions track actual clinical need~\citep{obermeyer2019dissecting,paulus2020predictably}. Standard parity checks could not have flagged the Obermeyer model, which satisfied within-group calibration while still under-scoring patients with the most unmet need.

\textbf{Third}, threshold tuning is a routine deployment step, yet no widely-used evaluation reports how much that tuning changes the patient flag set~\citep{wynants2020prediction}. \citet{macrae2024risk} analyses how autonomous and intelligent systems in healthcare can blur the sharp category boundaries that human decision-makers rely on for hazard response; RISED's Sensitivity dimension makes one facet of this concern---instability of the patient flag set under threshold shifts---directly measurable through the flip rate.

\textbf{Fourth}, operational feasibility---explanation consistency, inference speed at the bedside---is typically siloed in the engineering team while statistical performance is reported by the data science team~\citep{antoniadi2021xai,sutton2020overview}. For users who must act on model outputs in real time this separation is unhelpful and, in some patterns, actively unsafe~\citep{rudin2019stop,sendak2022model_facts}.

A reproducibility lens sharpens these gaps further: \citet{kapoor2023leakage} document that ML-based science systematically overstates performance through data leakage, held-out set contamination, and evaluation on non-representative samples. Standardised pre-deployment protocols with external-validation requirements are a structural response to this crisis; RISED's cohort-agnostic design and one-command reproducibility kit are designed with this concern in mind.

\paragraph{Positioning RISED against adjacent frameworks.}
No single existing tool covers the same ground as RISED.
\emph{Measurement toolkits} (AIF360~\citep{bellamy2019aif360}, Fairlearn~\citep{bird2020fairlearn,bird2023fairlearn}, Aequitas~\citep{saleiro2018aequitas}, FairLens~\citep{panigutti2021fairlens}) quantify subgroup-parity metrics but lack formal threshold-bearing tests, bootstrap-CI decision rules, and multi-dimension scope; CheckList~\citep{ribeiro2020checklist} packages behavioural testing for NLP but not tabular clinical data.
\emph{Production-readiness rubrics} (ML Test Score~\citep{breck2017mltestscore}) add infrastructure, pipeline health, and serving-latency tests; they omit statistical fairness, need-based equity, and threshold-sensitivity dimensions. \emph{ML lifecycle frameworks}~\citep{crespi2025lifecycle} model development end-to-end and identify explainability and governance gaps, but produce qualitative recommendations rather than threshold-bearing, bootstrap-CI-backed quantitative verdicts.
\emph{Disclosure artefacts} (model cards~\citep{mitchell2019modelcards}, datasheets~\citep{gebru2021datasheets}, DOME~\citep{walsh2021dome}, FAIR Principles~\citep{wilkinson2016fair}) structure \emph{what to report} rather than \emph{what numerical bar to clear}.
\emph{Organisational audit frameworks}~\citep{raji2020accountability} define governance process but explicitly name the absence of standardised quantitative test batteries as their primary gap.
\emph{Governance frameworks} (NIST AI RMF~\citep{nist2023aiirmf}, EU AI Act~\citep{euaiact2024}) mandate quantitative Measure-function evidence without specifying metrics or thresholds.
RISED combines threshold-bearing, BCa-bootstrapped, Holm--Bonferroni-corrected quantitative tests across five orthogonal deployment-relevant dimensions---an integration none of the adjacent tools provides---producing PASS/FAIL/INCONCLUSIVE verdicts that feed directly into the governance, disclosure, and toolkit layers without duplicating any.

Table~\ref{tab:tool_comparison} summarises how RISED differs from the most-used adjacent toolkits and protocols across the five dimensions.

\begin{table}[!htbp]
\centering
\caption{Coverage of pre-deployment evaluation concerns across the
most-used AI evaluation tools and protocols. \checkmark{} = packaged,
quantitative test; \textcircled{$\circ$}{} = partial / qualitative
coverage; --- = not in scope.
\textbf{Takeaway:} no existing tool offers a packaged quantitative test on all
five dimensions; RISED is the only complete row.}
\label{tab:tool_comparison}
\small
\begin{tabular}{lccccc}
\toprule
& Reliab. & Inclus. & Sensit. & Equity & Deploy. \\
& (encoding & (subgroup & (threshold & (need-based & (XAI + \\
& stability) & parity) & shifts) & alignment) & latency) \\
\midrule
AIF360~\citep{bellamy2019aif360}                         & ---                       & \checkmark                & ---                       & \textcircled{$\circ$}     & ---                       \\
Fairlearn~\citep{bird2020fairlearn,bird2023fairlearn}    & ---                       & \checkmark                & ---                       & \textcircled{$\circ$}     & ---                       \\
Aequitas~\citep{saleiro2018aequitas}                     & ---                       & \checkmark                & \textcircled{$\circ$}     & \textcircled{$\circ$}     & ---                       \\
FairLens~\citep{panigutti2021fairlens}                   & ---                       & \checkmark                & ---                       & \textcircled{$\circ$}     & ---                       \\
CheckList~\citep{ribeiro2020checklist}                   & \checkmark                & \textcircled{$\circ$}     & ---                       & ---                       & ---                       \\
ML Test Score~\citep{breck2017mltestscore}               & \textcircled{$\circ$}     & ---                       & ---                       & ---                       & \textcircled{$\circ$}     \\
TRIPOD+AI~\citep{collins2024tripodai}                    & \textcircled{$\circ$}     & \textcircled{$\circ$}     & \textcircled{$\circ$}     & \textcircled{$\circ$}     & \textcircled{$\circ$}     \\
APPRAISE-AI~\citep{kwong2023appraiseai}                  & \textcircled{$\circ$}     & \textcircled{$\circ$}     & ---                       & ---                       & \textcircled{$\circ$}     \\
FUTURE-AI~\citep{lekadir2025futureai}                    & \textcircled{$\circ$}     & \textcircled{$\circ$}     & ---                       & \textcircled{$\circ$}     & \textcircled{$\circ$}     \\
PROBAST~\citep{wolff2019probast}                         & \textcircled{$\circ$}     & ---                       & ---                       & ---                       & ---                       \\
\textbf{RISED} (this paper)                              & \textbf{\checkmark}       & \textbf{\checkmark}       & \textbf{\checkmark}       & \textbf{\checkmark}       & \textbf{\checkmark}       \\
\bottomrule
\end{tabular}
\end{table}

Table~\ref{tab:tool_comparison} reveals two structural gaps. On the \emph{coverage axis}, fairness toolkits (AIF360, Fairlearn, Aequitas) package Inclusivity; FairLens~\citep{panigutti2021fairlens} extends subgroup auditing to black-box CDSSs but covers Inclusivity only; behavioural testing packages Reliability for NLP; the ML Test Score~\citep{breck2017mltestscore} covers production infrastructure and partial Deployability but omits all four statistical dimensions; APPRAISE-AI~\citep{kwong2023appraiseai} provides qualitative grading across Reliability, Inclusivity, and Deployability but no quantitative threshold-bearing tests; reporting standards (TRIPOD+AI, FUTURE-AI) prescribe disclosure but do not implement numerical verdicts. On the \emph{decision-rule axis}, none combine a bootstrap-CI-based PASS / FAIL / INCONCLUSIVE verdict with multiple-testing correction across the full dimension set. To our knowledge, RISED is the first evaluation protocol packaging threshold-bearing quantitative tests across all five columns under Holm--Bonferroni FWER correction (\S\ref{sec:framework:thresholds}).

\section{The RISED Framework}
\label{sec:framework}

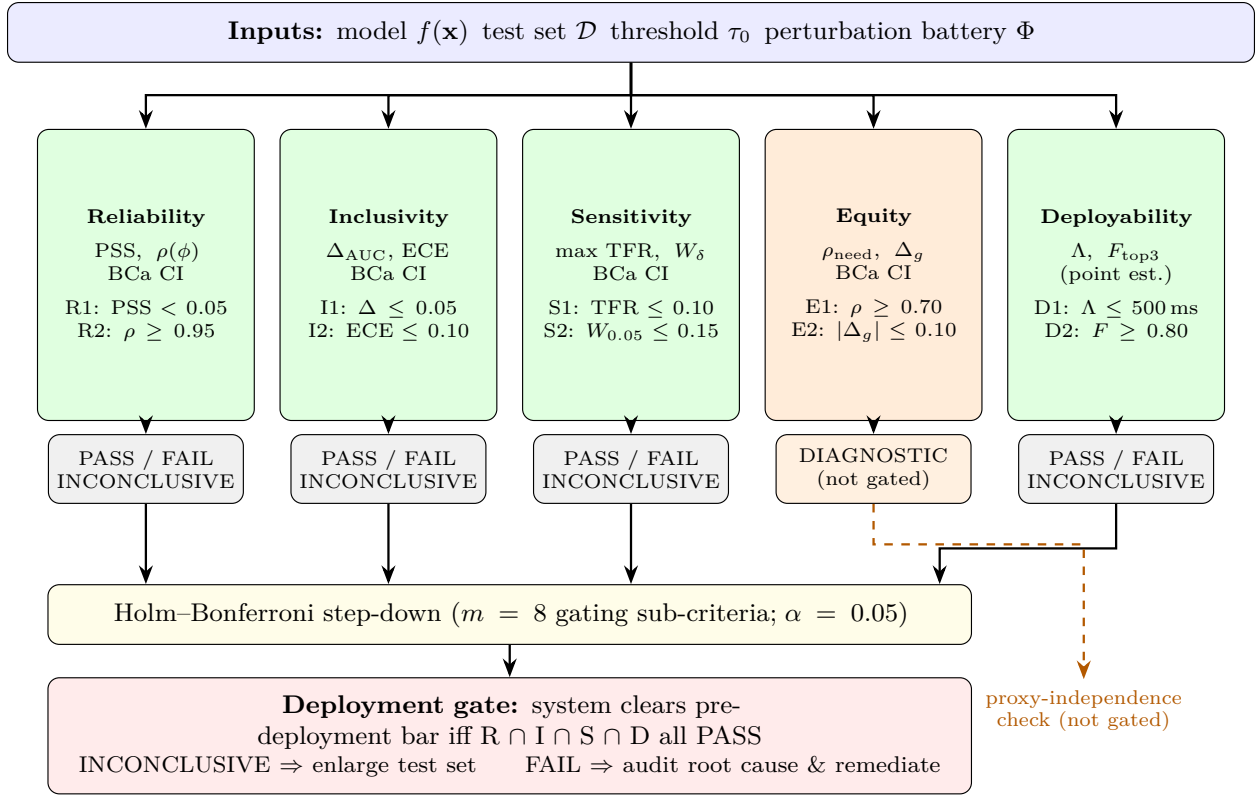
\begin{figure}[!t]
\centering
\resizebox{\linewidth}{!}{%
\begin{tikzpicture}[x=1cm,y=1cm,
  >=Stealth,
  arr/.style={->,thick},
  darr/.style={->,thick,dashed,orange!70!black},
  topbox/.style={draw,fill=blue!8,rounded corners=4pt,
                 align=center,font=\small,inner sep=6pt},
  gbox/.style={draw,fill=green!12,rounded corners=4pt,
               align=center,font=\scriptsize,inner sep=4pt,
               minimum width=2.6cm,minimum height=3.6cm,text width=2.4cm},
  ebox/.style={draw,fill=orange!15,rounded corners=4pt,
               align=center,font=\scriptsize,inner sep=4pt,
               minimum width=2.6cm,minimum height=3.6cm,text width=2.4cm},
  vbox/.style={draw,fill=gray!12,rounded corners=4pt,
               align=center,font=\scriptsize,inner sep=3pt,
               minimum width=2.4cm,minimum height=0.85cm,text width=2.2cm},
  evbox/.style={draw,fill=orange!12,rounded corners=4pt,
                align=center,font=\scriptsize,inner sep=3pt,
                minimum width=2.4cm,minimum height=0.85cm,text width=2.2cm},
  hbox/.style={draw,fill=yellow!10,rounded corners=4pt,
               align=center,font=\small,inner sep=6pt,text width=11cm},
  fbox/.style={draw,fill=red!8,rounded corners=4pt,
               align=center,font=\small,inner sep=6pt,text width=11cm},
]

\node[topbox,text width=15cm] (inp) at (0,0) {%
  \textbf{Inputs:}\enspace model $f(\mathbf{x})$\enspace
  test set $\mathcal{D}$\enspace
  threshold $\tau_0$\enspace
  perturbation battery $\Phi$};

\node[gbox] (R) at (-6,-3) {%
  \textbf{Reliability}\\[4pt]
  PSS,\; $\rho(\phi)$\\BCa CI\\[4pt]
  R1: PSS $< 0.05$\\R2: $\rho \ge 0.95$};

\node[gbox] (I) at (-3,-3) {%
  \textbf{Inclusivity}\\[4pt]
  $\Delta_\mathrm{AUC}$, ECE\\BCa CI\\[4pt]
  I1: $\Delta \le 0.05$\\I2: ECE $\le 0.10$};

\node[gbox] (S) at (0,-3) {%
  \textbf{Sensitivity}\\[4pt]
  max TFR,\; $W_\delta$\\BCa CI\\[4pt]
  S1: TFR $\le 0.10$\\S2: $W_{0.05} \le 0.15$};

\node[ebox] (E) at (3,-3) {%
  \textbf{Equity}\\[4pt]
  $\rho_\text{need}$,\; $\Delta_g$\\BCa CI\\[4pt]
  E1: $\rho \ge 0.70$\\E2: $|\Delta_g| \le 0.10$};

\node[gbox] (D) at (6,-3) {%
  \textbf{Deployability}\\[4pt]
  $\Lambda$,\; $F_\text{top3}$\\(point est.)\\[4pt]
  D1: $\Lambda \le 500$\,ms\\D2: $F \ge 0.80$};

\node[vbox]  (Rv) at (-6,-5.4) {PASS / FAIL\\INCONCLUSIVE};
\node[vbox]  (Iv) at (-3,-5.4) {PASS / FAIL\\INCONCLUSIVE};
\node[vbox]  (Sv) at ( 0,-5.4) {PASS / FAIL\\INCONCLUSIVE};
\node[evbox] (Ev) at ( 3,-5.4) {DIAGNOSTIC\\(not gated)};
\node[vbox]  (Dv) at ( 6,-5.4) {PASS / FAIL\\INCONCLUSIVE};

\draw[arr] (inp.south) -- ++(0,-0.4) -| (R.north);
\draw[arr] (inp.south) -- ++(0,-0.4) -| (I.north);
\draw[arr] (inp.south) -- ++(0,-0.4)  -- (S.north);
\draw[arr] (inp.south) -- ++(0,-0.4) -| (E.north);
\draw[arr] (inp.south) -- ++(0,-0.4) -| (D.north);

\draw[arr] (R.south) -- (Rv.north);
\draw[arr] (I.south) -- (Iv.north);
\draw[arr] (S.south) -- (Sv.north);
\draw[arr] (E.south) -- (Ev.north);
\draw[arr] (D.south) -- (Dv.north);

\node[hbox] (holm) at (-1.5,-7.2) {%
  Holm--Bonferroni step-down ($m=8$ gating sub-criteria;\;$\alpha=0.05$)};

\draw[arr] (Rv.south) -- (Rv.south |- holm.north);
\draw[arr] (Iv.south) -- (Iv.south |- holm.north);
\draw[arr] (Sv.south) -- (Sv.south |- holm.north);
\draw[arr] (Dv.south) -- ++(0,-0.55) -| ([xshift=-4mm]holm.north east);

\draw[darr] (Ev.south) -- ++(0,-0.5) -| (5.6,-8.0)
  node[below,font=\scriptsize,text=orange!70!black,align=center]
  {proxy-independence\\check (not gated)};

\node[fbox] (gate) at (-1.5,-8.7) {%
  \textbf{Deployment gate:} system clears pre-deployment bar iff
  R $\cap$ I $\cap$ S $\cap$ D all PASS\\[-1pt]
  {\footnotesize INCONCLUSIVE $\Rightarrow$ enlarge test set\qquad
   FAIL $\Rightarrow$ audit root cause \& remediate}};

\draw[arr] (holm.south) -- (gate.north);

\end{tikzpicture}}
\caption{RISED evaluation pipeline. Five parallel paths evaluate the model
  against eight gating sub-criteria (green boxes) and two Equity diagnostic
  sub-criteria (orange). BCa bootstrap 95\% CIs and Holm--Bonferroni FWER
  correction ($m=8$, $\alpha=0.05$) produce PASS / FAIL / INCONCLUSIVE verdicts
  for the four gating dimensions; Equity yields a DIAGNOSTIC verdict excluded
  from the deployment gate. An INCONCLUSIVE verdict signals that the test set
  is below the informative-verdict floor (\S\ref{sec:framework:thresholds});
  a FAIL triggers root-cause investigation before deployment is approved.
  \textbf{Takeaway:} all five dimensions run through one code path; four gate
  deployment by conjunction while Equity is diagnostic-only.}
\label{fig:pipeline}
\end{figure}

Figure~\ref{fig:pipeline} shows the end-to-end evaluation pipeline.
Given a trained model $f(\mathbf{x})$, a held-out test set $\mathcal{D}$,
an operating threshold $\tau_0$, and a perturbation battery $\Phi$, RISED
computes eight gating sub-criteria across four dimensions
(\textbf{R}eliability: R1, R2; \textbf{I}nclusivity: I1, I2;
\textbf{S}ensitivity: S1, S2; \textbf{D}eployability: D1, D2)
and two diagnostic sub-criteria for the non-gating
\textbf{E}quity dimension (E1, E2).
Each gating sub-criterion yields a bootstrap 95\% BCa CI and a
one-sided $p$-value; Holm--Bonferroni step-down at $m=8$ controls
the family-wise error rate, and the CI-based rule
(\S\ref{sec:framework:thresholds}) produces a
PASS / FAIL / INCONCLUSIVE verdict.
All five dimensions pass through the same code path; Equity is
distinguished only by being excluded from the gating conjunction.
The five sub-sections below specify each dimension's formal
sub-criteria, literature justification for the default threshold,
and scope boundaries.

\subsection{Dimension 1: Reliability}
\label{sec:framework:reliability}

A model may encounter inputs that are semantically equivalent yet encoded differently: the same diagnosis at a different ICD granularity, the same lab value in different units, or the same comorbidity at a slightly different date. When such encodings change model outputs, patients with identical clinical states are prioritised based on administrative artefacts rather than medical need. A scoping review catalogued eight robustness notions in healthcare ML, with input-perturbation stability among the least tested~\citep{balendran2025robustness}. Reliability applies a battery of semantically preserving perturbations and measures how often decisions and rankings change.

A \emph{perturbation} $\phi$ maps an input $\mathbf{x}$ to a semantically equivalent variant $\tilde{\mathbf{x}} = \phi(\mathbf{x})$ preserving the patient's clinical state. The \textbf{Perturbation Flip Rate (PFR)} is the fraction of patients whose binary decision changes under $\phi$ at threshold $\tau$:
\begin{equation}
  \mathrm{PFR}(\phi, \tau) =
    \frac{1}{n}\sum_{i=1}^{n}
    \mathbf{1}\!\left[
      \mathbf{1}[f(\mathbf{x}_i) \ge \tau] \ne
      \mathbf{1}[f(\phi(\mathbf{x}_i)) \ge \tau]
    \right].
  \label{eq:pfr}
\end{equation}
A PFR above 0.05 means one patient in twenty would be classified differently by a trivial encoding change. Averaging PFR across battery $\Phi$ yields the \textbf{Perturbation Sensitivity Score (PSS)}:
\begin{equation}
  \mathrm{PSS}(\Phi, \tau) = \frac{1}{|\Phi|} \sum_{\phi \in \Phi}
    \mathrm{PFR}(\phi, \tau).
  \label{eq:pss}
\end{equation}
PSS is a single-number summary of perturbation-induced decision instability, in the same family as the robustness metrics of~\citet{balendran2025robustness} and~\citet{subbaswamy2021development}, instantiated for clinically realistic encoding shifts rather than $\ell_p$-bounded adversarial perturbations. We complement PSS with the Spearman rank correlation $\rho(\phi)$ between baseline and perturbed score vectors, capturing ordering stability independent of the binary threshold.

\medskip
\noindent\textbf{Sub-criteria.}
\begin{itemize}[leftmargin=2em]
  \item[\textbf{R1}] $\mathrm{PSS}(\Phi, \tau) < 0.05$: fewer than 5\% of patients
    receive a flipped binary decision on average across the perturbation battery.
  \item[\textbf{R2}] $\rho(\phi) \ge 0.95$ for all $\phi \in \Phi$: the relative
    ordering of patients is preserved across every perturbation variant.
\end{itemize}

\paragraph{Threat model and scope of the PSS metric.}
PSS is not a certified-robustness or worst-case adversarial
guarantee~\citep{madry2018adversarial}; it does not bound model behaviour
over an $\ell_p$ ball. Rather, PSS reports flip rate over a user-specified
\emph{ensemble} of semantically motivated perturbations, and is therefore a
\emph{property of the chosen perturbation battery}, not a model invariant: a
different battery yields a different PSS, and reviewers should evaluate the
two together. The package documentation specifies how to construct batteries
approximating real deployment-time coding transitions (ICD-9$\to$ICD-10,
LOINC harmonisation, unit changes mg/dL$\leftrightarrow$mmol/L); isotropic
Gaussian noise is reported in this paper as a baseline, not as a final
clinical battery. Data standardisation is a prerequisite: PSS values are only comparable across sites if the perturbation battery reflects real encoding transitions at the target site, and multi-site deployments should extend the battery with site-specific ICD-version, LOINC, and unit-conversion profiles.

\subsection{Dimension 2: Inclusivity}
\label{sec:framework:inclusivity}

A model may achieve strong aggregate discrimination while performing markedly worse for specific patient subpopulations. Race, sex, and age are the most consistently documented axes of clinical outcome disparity~\citep{obermeyer2019dissecting,celi2022sources,osibogun2024adverse} and are also those the FDA AI/ML Action Plan~\citep{fda2021aiml} and ONC HTI-1 rule~\citep{onc2024hti1} require reporting on. We note that race in clinical algorithms is a fraught construct: \citet{vyas2020hidden} document at least thirteen widely deployed clinical algorithms (eGFR, vaginal-birth-after-cesarean, STONE) whose race-correction terms produced demonstrable patient harm. RISED's Inclusivity dimension surfaces subgroup-AUC and subgroup-calibration gaps; whether such a gap reflects a genuine biological signal or a structural-inequity artefact remains a domain-expert judgment that the framework cannot make unilaterally. Insurance type is added because access-coverage status encodes utilisation-based disparity~\citep{paulus2020predictably}; other partitions can be supplied via the package's user-defined subgroup keys. Beyond discrimination, calibration must hold within subgroups: systematic probability over- or underestimation for a particular group causes incorrect decisions even when aggregate calibration looks clean~\citep{pleiss2017fairness,van_calster2019calibration}. The Inclusivity dimension jointly evaluates subgroup discrimination and calibration.

Let $\mathcal{G}$ denote a set of subgroups partitioning patients by a demographic
attribute, and let $\mathrm{AUC}_g$ denote the ROC-AUC computed on patients in subgroup
$g$ only. The \textbf{AUC Parity Gap} captures the worst-case discrimination disparity
across subgroups:
\begin{equation}
  \Delta_\mathrm{AUC}(\mathcal{G}) =
    \max_{g \in \mathcal{G}}\,\mathrm{AUC}_g
    - \min_{g \in \mathcal{G}}\,\mathrm{AUC}_g.
  \label{eq:auc-gap}
\end{equation}
Subgroup calibration is assessed via the \textbf{Expected Calibration Error (ECE)} within each subgroup~\citep{guo2017calibration,van_calster2019calibration}, using equal-width probability bins:
\begin{equation}
  \mathrm{ECE}_g = \sum_{b=1}^{B}
    \frac{|\mathcal{B}_{g,b}|}{|\mathcal{I}_g|}
    \left|\bar{y}_{g,b} - \bar{f}_{g,b}\right|,
  \label{eq:ece}
\end{equation}
where $\mathcal{B}_{g,b}$ is the $b$-th probability bin within subgroup $g$, $\bar{y}_{g,b}$ is the mean observed label, and $\bar{f}_{g,b}$ is the mean predicted probability.

\medskip
\noindent\textbf{Sub-criteria.}
\begin{itemize}[leftmargin=2em]
  \item[\textbf{I1}] $\Delta_\mathrm{AUC} \le 0.05$ per demographic partition: the
    worst-performing subgroup achieves AUC within 5 percentage points of the
    best-performing subgroup.
  \item[\textbf{I2}] $\mathrm{ECE}_g \le 0.10$ for all subgroups $g$: no subgroup suffers
    systematic probability miscalibration exceeding 10 percentage points.
  \item[\textbf{I3}] Subgroups with fewer than 30 patients are flagged for informational
    purposes; their metrics are reported but do not count toward pass/fail, as sample sizes
    are insufficient to estimate AUC reliably~\citep{steyerberg2010assessing}.
\end{itemize}

\subsection{Dimension 3: Sensitivity}
\label{sec:framework:sensitivity}

The decision threshold is routinely adjusted post-deployment, yet standard evaluation reports performance at one fixed threshold. The Sensitivity dimension measures, for a reference threshold $\tau_0$, how many patients change classification when the threshold moves to $\tau$.

For a predicted score vector $f(\mathbf{X}) \in [0,1]^n$, the \textbf{Threshold Flip
Rate} at threshold $\tau$ relative to reference $\tau_0$ is
\begin{equation}
  \mathrm{TFR}(\tau, \tau_0) = \frac{1}{n}\sum_{i=1}^{n}
    \mathbf{1}\!\left[
      \mathbf{1}[f(\mathbf{x}_i) \ge \tau_0] \ne
      \mathbf{1}[f(\mathbf{x}_i) \ge \tau]
    \right].
  \label{eq:tfr}
\end{equation}
Evaluating TFR across sweep $\Theta = \{\tau_1, \ldots, \tau_K\}$ (default $K=17$ thresholds in $[0.10, 0.90]$) characterises the full sensitivity profile; the package also reports max TFR over the clinically-realistic $[0.30, 0.70]$ band. The \textbf{decision boundary width} $W_\delta$ quantifies the fraction of patients whose score lies within $\delta$ of $\tau_0$:
\begin{equation}
  W_\delta(\tau_0) = \frac{1}{n}\sum_{i=1}^{n}
    \mathbf{1}\!\left[|f(\mathbf{x}_i) - \tau_0| \le \delta\right].
  \label{eq:boundary-width}
\end{equation}
A large $W_\delta$ indicates many patients are near-borderline.

\medskip
\noindent\textbf{Sub-criteria.}
\begin{itemize}[leftmargin=2em]
  \item[\textbf{S1}] $\max_{\tau \in \Theta}\,\mathrm{TFR}(\tau, \tau_0) \le 0.10$ across the evaluation sweep $\Theta$ (default: 17 thresholds in $[0.10, 0.90]$).
  \item[\textbf{S2}] $W_{0.05}(\tau_0) \le 0.15$: at most 15\% of patients are
    borderline-sensitive to threshold perturbations of $\pm 5$ percentage points.
\end{itemize}

\subsection{Dimension 4: Equity}
\label{sec:framework:equity}

Standard fairness criteria evaluate the relationship between model predictions and observed outcomes, typically healthcare utilisation or cost. As \citet{paulus2020predictably} show, this breaks down when utilisation is itself distorted by access barriers: a model can satisfy within-group calibration while still under-scoring the groups with the greatest unmet need. Equity operationalises \emph{need-based fairness}: alignment of predictions with independent measures of clinical need.

Let $s_i = f(\mathbf{x}_i)$ be the predicted score and $c_i$ a clinical need measure (e.g.\ a normalised comorbidity count). The \textbf{need-prediction correlation} measures global alignment:
\begin{equation}
  \rho_\text{need} = \mathrm{Spearman}(s,\, c).
  \label{eq:need-corr}
\end{equation}
The \textbf{group need gap} for subgroup $g$ captures directional misalignment:
\begin{equation}
  \Delta_g = \bar{s}_g - \bar{c}_g,
  \label{eq:group-gap}
\end{equation}
where $\bar{s}_g$ and $\bar{c}_g$ are the mean predicted score and mean clinical need in group $g$. A negative $\Delta_g$ indicates the model \emph{under-predicts} need for group $g$.

\medskip
\noindent\textbf{Sub-criteria.}
\begin{itemize}[leftmargin=2em]
  \item[\textbf{E1}] $\rho_\text{need} \ge 0.70$: the model's scores rank patients
    by clinical need with at least moderate Spearman rank correlation.
  \item[\textbf{E2}] $|\Delta_g| \le 0.10$ for all subgroups $g$: no subgroup is
    systematically under- or over-scored by more than 10 percentage points relative
    to its mean clinical need.
\end{itemize}

\paragraph{Choosing a need proxy.}
A valid Equity audit requires a need proxy that satisfies three
conditions: \emph{(i)~outcome-independence}, the proxy must not be a
function of the training label nor of features used to construct
that label; \emph{(ii)~clinical face validity}, the proxy must be
defensible by a clinician as tracking the underlying construct the model
is meant to prioritise---clinical need, not utilisation; and
\emph{(iii)~availability outside the training pipeline}, the proxy must
come from a measurement upstream or orthogonal to the feature pipeline
used by the model. Clinical examples that satisfy all three: prospectively
recorded nurse acuity scores, triage severity levels, and post-discharge
mortality linked externally. Examples that do \emph{not} satisfy them
and are common mistakes: the training label itself, a feature already
used in training, or a downstream utilisation variable that encodes the
same access barriers being audited. When no proxy in the available data passes all three tests,
the appropriate Equity verdict is DIAGNOSTIC, not FAIL---the audit
is informative-only until a defensible proxy is procured.

\paragraph{Diagnostic framing.}
Because $\rho_\text{need}$ depends on the chosen proxy, the framework treats Equity as a \emph{diagnostic of proxy-dependence}: a verdict change between proxies triggers procurement of an outcome-independent need measure, not an automatic deployment stop. The \texttt{rised} package emits a \texttt{UserWarning} when the outcome label is used as the proxy, and Equity is excluded from the all-pass gating conjunction regardless of the underlying $\rho_\mathrm{need}$ value.

\subsection{Dimension 5: Deployability}
\label{sec:framework:deployability}

A model that clears the statistical bar may still fail the workflow bar: two-second latency does not fit a dashboard refresh, and explanations highlighting different features for nearly-identical patients cannot support consistent clinical judgment~\citep{sutton2020overview,antoniadi2021xai,rudin2019stop}. Deployability targets these operational properties in settings where end users are not ML specialists.

\textbf{Inference latency} is the mean wall-clock time over $R$ calls:
\begin{equation}
  \Lambda = \frac{1}{R}\sum_{r=1}^{R} t_r,
  \label{eq:latency}
\end{equation}
where $t_r$ is the time in milliseconds for the $r$-th call. \textbf{Explanation top-3 consistency} ($F_{\text{top3}}$) measures whether each patient's locally most important feature (by SHAP magnitude~\citep{lundberg2017shap}) is among the three globally most important features:
\begin{equation}
  F_\text{top3} = \frac{1}{n}\sum_{i=1}^{n}
    \mathbf{1}\!\left[
      \arg\max_j\,|\phi_{ij}| \in \mathcal{T}_3
    \right],
  \label{eq:top3consistency}
\end{equation}
where $\phi_{ij}$ is the SHAP value for patient $i$, feature $j$, and $\mathcal{T}_3$ is the set of the three globally most important features by mean $|\phi|$. Note: $F_\text{top3}$ measures self-consistency of SHAP attributions between global and local views, not faithfulness to the decision boundary in the \citet{jacovi2020faithfulness} sense. A high $F_\text{top3}$ means the features a clinician associates with high-risk predictions also drive individual scores, supporting consistent interpretation~\citep{sendak2022model_facts,tonekaboni2019whatclinicians}.

\medskip
\noindent\textbf{Sub-criteria.}
\begin{itemize}[leftmargin=2em]
  \item[\textbf{D1}] $\Lambda \le 500\,\text{ms}$ per cohort: the model processes a
    full patient batch within a real-time operational limit compatible with dashboard
    refresh requirements.
  \item[\textbf{D2}] $F_\text{top3} \ge 0.80$: globally important features are
    locally relevant for at least 80\% of patients, supporting consistent
    clinician interpretation across individual predictions.
\end{itemize}

\noindent D1 and D2 assess model properties only. Whether a HIPAA-compliant inference pipeline---encrypted transit, patient-level de-identification, audit logging---can be constructed for a given EHR environment is an infrastructure question outside RISED's scope.

\subsection{Default Thresholds and Decision Rule}
\label{sec:framework:thresholds}

Default thresholds (Table~\ref{tab:thresholds}) are literature-grounded starting values, not empirically calibrated constants; all are user-configurable and should be recalibrated for the target use case. The framework operates with \emph{four gating dimensions} (Reliability, Inclusivity, Sensitivity, Deployability); Equity is a proxy-dependence diagnostic excluded from the gating conjunction (\S\ref{sec:framework:equity}). Headline CIs use BCa bootstrap~\citep{efron1987better,davison1997bootstrap} with $B=1{,}000$ resamples (random state 42); we recommend $B\ge 2{,}000$ when CI endpoints sit near a threshold. FWER across $m=8$ gating sub-criteria is controlled by Holm--Bonferroni step-down. Statistical justification, power analysis, and hypothesis-framing details are in Appendix~\ref{app:statistics}.

\paragraph{CI-based decision rule.}
For sub-criteria with a bootstrap 95\% CI (PSS, $\Delta_\mathrm{AUC}$, max TFR):
\begin{itemize}[leftmargin=2em,topsep=2pt]
  \item \textbf{PASS} if the entire 95\% BCa CI lies in the accept region:
        for upper-bounded sub-criteria (R1: PSS$<0.05$, I1:
        $\Delta_\mathrm{AUC}\le 0.05$, S1: max TFR $\le 0.10$), the CI
        \emph{upper bound} is below the threshold; for lower-bounded
        sub-criteria (E1: $\rho_\mathrm{need}\ge 0.70$), the CI \emph{lower
        bound} is above the threshold;
  \item \textbf{FAIL} if the entire 95\% BCa CI lies in the reject region
        (the symmetric reverse);
  \item \textbf{INCONCLUSIVE} otherwise (CI brackets the threshold); signals that a larger test set is needed;
  \item \textbf{DIAGNOSTIC} (Equity only): never combined into the gating conjunction; triggers a procurement requirement, not a deployment block (\S\ref{sec:framework:equity}).
\end{itemize}

\paragraph{Threshold sensitivity and metric monotonicity.}
We check verdict robustness in two ways. \emph{(a) Threshold sweep:} the Reliability verdict is FAIL at thresholds 0.025--0.050, and PASS at 0.075 and 0.10 (the CI upper bound of 0.070 falls below both, so no INCONCLUSIVE arises at those sweep points). Sensitivity is FAIL across the full 5\%--15\% band. \emph{(b) Monotonicity:} under Gaussian noise $\sigma\in\{0,\,0.025,\,0.05,\,0.10\}$, PSS increases near-monotonically from 0\% to ${\sim}10\%$, confirming the metric captures input sensitivity in the expected direction. The sweep script is \texttt{examples/threshold\_sensitivity.py}.

\begin{table}[!t]
\centering
\caption{RISED default pass/fail thresholds. These defaults are
informed by published clinical conventions where applicable (column 3),
but are not strictly derived constants and should be recalibrated
empirically for the target use case.
\textbf{Takeaway:} every threshold is a literature-grounded, user-configurable
starting value---not a hard-coded constant.}
\label{tab:thresholds}
\small
\begin{tabular}{llp{7.2cm}}
\toprule
\textbf{Dimension} & \textbf{Default} & \textbf{Basis (informative, not derivational)} \\
\midrule
Reliability   & PSS $<$ 0.05
  & 5\% is a common decision-error tolerance in clinical prediction
    model validation \citep{steyerberg2010assessing}; we adopt it as
    a sensible starting threshold for input-perturbation stability.
    Actual acceptable instability depends on deployment context. \\[1pt]
Inclusivity   & $\Delta_\mathrm{AUC} \le 0.05$
  & 5 percentage-point AUC parity gap reflects the spirit of FDA
    AI/ML Action Plan expectations for performance by demographic
    subgroup \citep{fda2021aiml} and is consistent with operational
    defaults used in fairness toolkits \citep{bellamy2019aif360}. \\[1pt]
Sensitivity   & max TFR $\le 0.10$
  & A threshold-shift-induced reclassification of more than 10\%
    constitutes a substantial change in flagged patients;
    Wynants et al.\ \citep{wynants2019thresholds} discuss the
    non-arbitrary nature of threshold selection. \\[1pt]
Equity        & $\rho_\mathrm{need} \ge 0.70$
  & Spearman $\rho \ge 0.70$ corresponds to a strong monotone
    relationship in conventional effect-size language
    \citep{cohen1988statistical}.
    Obermeyer et al.\ \citep{obermeyer2019dissecting} document
    real-world need-prediction misalignment as a clinically
    consequential failure mode. \\[1pt]
Deployability & $\Lambda \le 500\,\text{ms}$ (cohort) /
                $\le 1\,\text{ms}$ (per patient)
  & Default reflects an interactive-tool refresh
    constraint~\citep{sutton2020overview}. The default is generous;
    tighter per-patient thresholds (e.g., 1\,ms) may be appropriate
    for time-critical clinical contexts. The package reports both
    cohort and per-patient latencies. \\
\bottomrule
\end{tabular}
\end{table}

\section{Application: Synthetic Illustration and Six Real-Data Cohorts}
\label{sec:application}

RISED is applied to seven cohorts: a 10{,}000-patient synthetic cohort used to illustrate methodology and isolate the Equity proxy-circularity question, plus six real-data cohorts spanning 35 years of data vintage (UCI Heart Disease 1989, UCI Diabetes 130 1999--2008, NCHS NHIS 2024 cardiovascular, NCHS NHIS 2023 diabetes, NCHS NHANES 2021--2023 diabetes, CDC BRFSS 2024 CHD/MI). The synthetic AUROC carries no deployment-relevant signal; the real-data cohorts carry the substantive evidence. \S\ref{sec:application:data}--\S\ref{sec:application:results} cover the synthetic illustration; \S\ref{sec:application:realdata} reports the six real-data scorecards; \S\ref{sec:application:multimodel}--\S\ref{sec:application:tehai} report robustness, fairness-toolkit comparison, and framework coverage.

\subsection{Data: Synthetic Clinical Cohort}
\label{sec:application:data}

We generated a synthetic cohort of 10,000 patients using a Synthea-inspired generative model~\citep{walonoski2018synthea} (\texttt{rised.datasets}; no real patient records; openly released per the Data Availability statement). The 20-variable feature matrix covers demographics, 9 chronic-condition flags, CCI, utilisation counters, anthropometrics, and a neighbourhood deprivation index (Table~\ref{tab:cohort-summary}).

The binary outcome is a noisy logistic function of training features (top 30\% positive; $n_+=3{,}000$); AUROC 0.961 reflects the data-generating process, not model skill. Train/test split: $n_\text{train}=8{,}000$, $n_\text{test}=2{,}000$ (stratified, seed 42).

\begin{table}[ht]
\centering
\caption{Summary characteristics of the 10,000-patient synthetic cohort.
\textbf{Takeaway:} the synthetic cohort carries realistic demographic and clinical
variation, serving only to illustrate the method (clinical conclusions rest on the
six real cohorts).}
\label{tab:cohort-summary}
\begin{tabular}{lrr}
\toprule
\textbf{Characteristic} & \textbf{Full cohort ($n$=10,000)} & \textbf{Outcome$=$1 ($n$=3,000)} \\
\midrule
\multicolumn{3}{l}{\textit{Age group}} \\
\quad 18--44 & 1,835 (18.4\%) & 6 (0.2\%) \\
\quad 45--64 & 2,502 (25.0\%) & 277 (9.2\%) \\
\quad 65--74 & 2,822 (28.2\%) & 947 (31.6\%) \\
\quad 75+ & 2,841 (28.4\%) & 1,770 (59.0\%) \\
\multicolumn{3}{l}{\textit{Sex}} \\
\quad Female & 5,549 (55.5\%) & 1,675 (55.8\%) \\
\quad Male & 4,451 (44.5\%) & 1,325 (44.2\%) \\
\multicolumn{3}{l}{\textit{Race/ethnicity}} \\
\quad White & 6,379 (63.8\%) & 1,898 (63.3\%) \\
\quad Black & 1,343 (13.4\%) & 404 (13.5\%) \\
\quad Hispanic & 1,301 (13.0\%) & 393 (13.1\%) \\
\quad Asian & 569 (5.7\%) & 176 (5.9\%) \\
\quad Other & 408 (4.1\%) & 129 (4.3\%) \\
\multicolumn{3}{l}{\textit{Insurance}} \\
\quad Medicare & 4,737 (47.4\%) & 2,225 (74.2\%) \\
\quad Private & 2,977 (29.8\%) & 413 (13.8\%) \\
\quad Medicaid & 1,444 (14.4\%) & 230 (7.7\%) \\
\quad Uninsured & 842 (8.4\%) & 132 (4.4\%) \\
\multicolumn{3}{l}{\textit{Clinical measures (mean $\pm$ SD)}} \\
\quad CCI score & $0.99 \pm 1.20$ & $1.86 \pm 1.33$ \\
\quad BMI & $28.7 \pm 6.0$ & $28.6 \pm 6.0$ \\
\quad Deprivation index & $49.8 \pm 24.1$ & $51.1 \pm 24.1$ \\
\multicolumn{3}{l}{\textit{Outcome}} \\
\quad Adverse clinical event & 3,000 (30.0\%) & 3,000 (100\%) \\
\bottomrule
\end{tabular}
\end{table}

\subsection{Baseline Model}
\label{sec:application:model}

We trained an XGBoost classifier~\citep{chen2016xgboost} on the 8,000-patient split (200 rounds, max depth 4, lr 0.05, subsample 0.80, colsample 0.80). A fallback to scikit-learn's \texttt{HistGradientBoostingClassifier}~\citep{sklearn2011} is provided for environments without XGBoost. AUROC 0.961 and Brier score 0.073 are expected given the self-derived outcome; the purpose is to show RISED surfaces deployment risks invisible to aggregate metrics.

\subsection{RISED Evaluation Results}
\label{sec:application:results}

\begin{figure}[!t]
\centering
\resizebox{\linewidth}{!}{\begin{tikzpicture}
\begin{axis}[
  width=15cm, height=6.2cm,
  xbar, bar width=11pt,
  xmin=0, xmax=14, ymin=-0.7, ymax=3.7,
  ytick={0,1,2,3},
  yticklabels={Age rescale $+5\%$, Age rescale $+6\%$, Gaussian noise $\sigma{=}5\%$, Gaussian noise $\sigma{=}10\%$},
  xlabel={Decision flip rate (\%)},
  title={\textbf{Reliability: decision flip rate by perturbation type}},
  xmajorgrids, grid style={dotted,gray!50},
  axis y line*=left, axis x line*=bottom,
  enlarge y limits=false,
  legend style={at={(0.98,0.05)},anchor=south east,draw=gray!50,font=\footnotesize},
]
\addplot[forget plot,xbar,bar shift=0pt,fill=cpass,draw=cpass!50!black,
  nodes near coords,point meta=x,
  every node near coord/.append style={anchor=west,font=\small\bfseries,text=cpass,
    /pgf/number format/fixed,/pgf/number format/precision=1},
  nodes near coords={\pgfmathprintnumber\pgfplotspointmeta\%}]
  coordinates {(2.5,0) (3.4,1)};
\addplot[forget plot,xbar,bar shift=0pt,fill=cfail,draw=cfail!50!black,
  nodes near coords,point meta=x,
  every node near coord/.append style={anchor=west,font=\small\bfseries,text=cfail,
    /pgf/number format/fixed,/pgf/number format/precision=1},
  nodes near coords={\pgfmathprintnumber\pgfplotspointmeta\%}]
  coordinates {(9.8,2) (10.1,3)};
\fill[cincon,opacity=0.12] (axis cs:5.8,-0.7) rectangle (axis cs:7.0,3.7);
\draw[cthresh,thick,dashed] (axis cs:5,-0.7) -- (axis cs:5,3.7);
\draw[cincon,very thick,dotted] (axis cs:6.4,-0.7) -- (axis cs:6.4,3.7);
\addlegendimage{area legend,fill=cpass,draw=cpass!50!black}\addlegendentry{Below threshold (PASS)}
\addlegendimage{area legend,fill=cfail,draw=cfail!50!black}\addlegendentry{Exceeds threshold (FAIL)}
\addlegendimage{cthresh,dashed,thick}\addlegendentry{Pass threshold (5\%)}
\addlegendimage{cincon,very thick,dotted}\addlegendentry{PSS $=0.064$ [0.058, 0.070]}
\end{axis}
\end{tikzpicture}}
\caption{Synthetic cohort --- Reliability: decision flip rate per perturbation.
         Dashed line: 5\% pass threshold. Red bars (Gaussian noise) exceed it;
         green bars (age rescalings) pass. PSS $= 0.064$ [0.058, 0.070];
         CI above 0.05 $\Rightarrow$ FAIL.
         \textbf{Takeaway:} reliability is perturbation-type dependent---encoding
         noise flips $\sim$10\% of decisions while unit rescalings stay stable.}
\label{fig:reliability}
\end{figure}
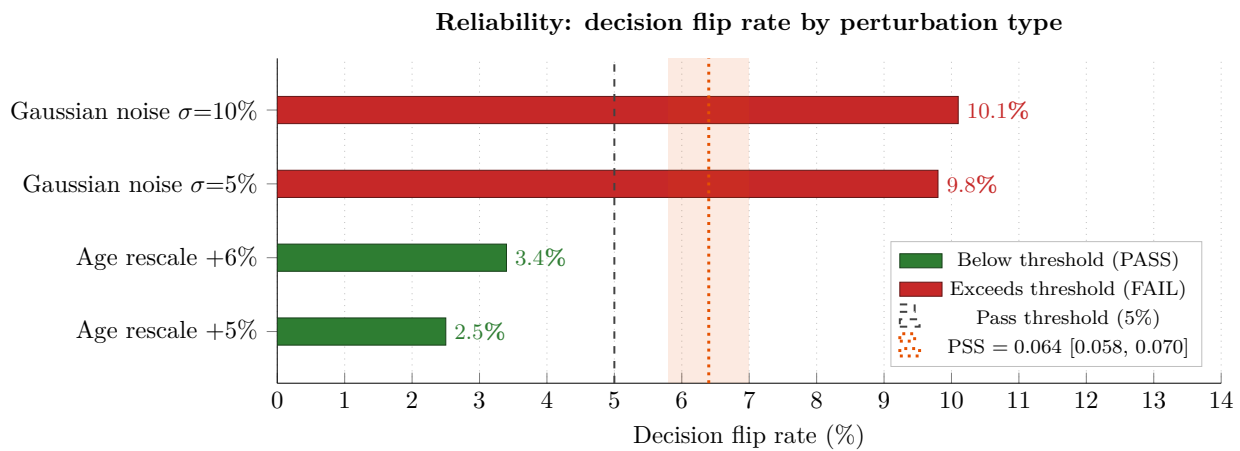

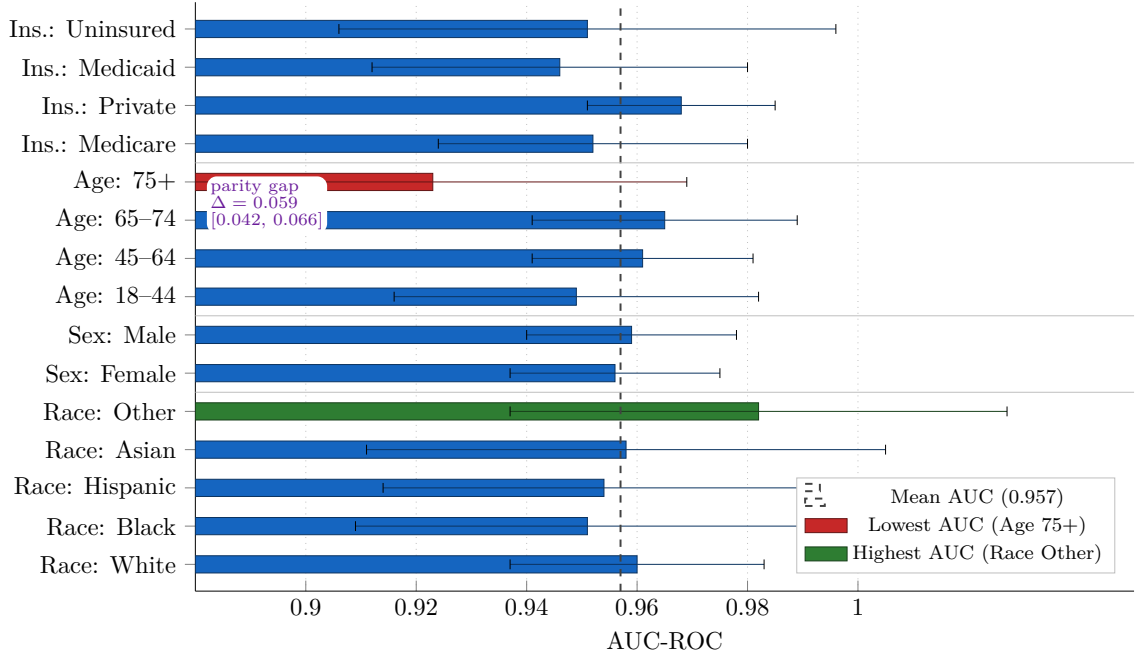
\begin{figure}[!t]
\centering
\resizebox{\linewidth}{!}{\begin{tikzpicture}
\begin{axis}[
  width=15cm, height=10cm,
  xbar, bar width=7pt,
  xmin=0.88, xmax=1.05, ymin=-0.7, ymax=14.7,
  ytick={0,1,2,3,4,5,6,7,8,9,10,11,12,13,14},
  yticklabels={Race: White,Race: Black,Race: Hispanic,Race: Asian,Race: Other,
    Sex: Female,Sex: Male,Age: 18--44,Age: 45--64,Age: 65--74,Age: 75+,
    Ins.: Medicare,Ins.: Private,Ins.: Medicaid,Ins.: Uninsured},
  xtick={0.90,0.92,0.94,0.96,0.98,1.00},
  xlabel={AUC-ROC},
  title={\textbf{Inclusivity: subgroup AUC-ROC with 95\% CIs (max parity gap $=0.059$; INCONCLUSIVE)}},
  xmajorgrids, grid style={dotted,gray!50},
  axis y line*=left, axis x line*=bottom,
  legend style={at={(0.98,0.03)},anchor=south east,draw=gray!50,font=\footnotesize},
]
\addplot[forget plot,xbar,bar shift=0pt,fill=cbar,draw=cbar!50!black,
  error bars/.cd,x dir=both,x explicit]
  coordinates {
   (0.960,0)+-(0.023,0) (0.951,1)+-(0.042,0) (0.954,2)+-(0.040,0)
   (0.958,3)+-(0.047,0) (0.956,5)+-(0.019,0) (0.959,6)+-(0.019,0)
   (0.949,7)+-(0.033,0) (0.961,8)+-(0.020,0) (0.965,9)+-(0.024,0)
   (0.952,11)+-(0.028,0) (0.968,12)+-(0.017,0) (0.946,13)+-(0.034,0)
   (0.951,14)+-(0.045,0)};
\addplot[forget plot,xbar,bar shift=0pt,fill=cpass,draw=cpass!50!black,
  error bars/.cd,x dir=both,x explicit]
  coordinates {(0.982,4)+-(0.045,0)};
\addplot[forget plot,xbar,bar shift=0pt,fill=cfail,draw=cfail!50!black,
  error bars/.cd,x dir=both,x explicit]
  coordinates {(0.923,10)+-(0.046,0)};
\draw[cthresh,thick,dashed] (axis cs:0.957,-0.7) -- (axis cs:0.957,14.7);
\draw[gray!50,thin] (axis cs:0.88,4.5) -- (axis cs:1.05,4.5);
\draw[gray!50,thin] (axis cs:0.88,6.5) -- (axis cs:1.05,6.5);
\draw[gray!50,thin] (axis cs:0.88,10.5) -- (axis cs:1.05,10.5);
\node[ctop3,font=\scriptsize,anchor=west,align=left,fill=white,inner sep=1.5pt,
  rounded corners] at (axis cs:0.882,9.4)
  {parity gap\\[-1pt]$\Delta=0.059$\\[-1pt][0.042, 0.066]};
\addlegendimage{cthresh,dashed,thick}\addlegendentry{Mean AUC (0.957)}
\addlegendimage{area legend,fill=cfail,draw=cfail!50!black}\addlegendentry{Lowest AUC (Age 75+)}
\addlegendimage{area legend,fill=cpass,draw=cpass!50!black}\addlegendentry{Highest AUC (Race Other)}
\end{axis}
\end{tikzpicture}}
\caption{Synthetic cohort --- Inclusivity: subgroup AUC-ROC. Age 75+ (0.923) is
         lowest; Race:Other (0.982) highest; dashed line: mean AUC 0.957.
         Parity gap $= 0.059$ [0.042, 0.066]; CI brackets 0.05 $\Rightarrow$ INCONCLUSIVE.
         \textbf{Takeaway:} the worst subgroup trails the best by 0.06 AUC, but the
         CI straddles the 0.05 bar---a larger cohort is needed to rule decisively.}
\label{fig:subgroup}
\end{figure}

\begin{figure}[!t]
\centering
\resizebox{\linewidth}{!}{\begin{tikzpicture}
\begin{axis}[
  width=15cm, height=7.2cm,
  xbar, bar width=10pt,
  xmin=-0.14, xmax=0.14, ymin=-0.7, ymax=10.7,
  scaled x ticks=false, xtick={-0.10,-0.05,0,0.05,0.10},
  xticklabel style={/pgf/number format/fixed,/pgf/number format/precision=2},
  ytick={0,1,2,3,4,5,6,7,8,9,10},
  yticklabels={Race: White,Race: Black,Race: Hispanic,Race: Asian,Race: Other,
    Sex: Female,Sex: Male,Age: 18--44,Age: 45--64,Age: 65--74,Age: 75+},
  xlabel={Need--prediction gap (prediction score $-$ need proxy)},
  title={\textbf{Equity: group-level need--prediction gaps (outcome-label proxy)}},
  xmajorgrids, grid style={dotted,gray!50},
  axis y line*=left, axis x line*=bottom,
  legend style={at={(0.98,0.98)},anchor=north east,draw=gray!50,font=\footnotesize},
  nodes near coords,point meta=x,
  every node near coord/.append style={font=\footnotesize,
    /pgf/number format/fixed,/pgf/number format/precision=3,/pgf/number format/print sign},
]
\addplot[forget plot,xbar,bar shift=0pt,fill=cbar,draw=cbar!50!black,
  every node near coord/.append style={anchor=west}]
  coordinates {(0.011,0)(0.043,3)(0.019,4)(0.013,6)(0.072,7)(0.028,8)};
\addplot[forget plot,xbar,bar shift=0pt,fill=cbar,draw=cbar!50!black,
  every node near coord/.append style={anchor=east}]
  coordinates {(-0.048,1)(-0.031,2)(-0.012,5)(-0.015,9)(-0.062,10)};
\draw[black,thick] (axis cs:0,-0.7) -- (axis cs:0,10.7);
\draw[cthresh,thick,dashed] (axis cs:0.10,-0.7) -- (axis cs:0.10,10.7);
\draw[cthresh,thick,dashed] (axis cs:-0.10,-0.7) -- (axis cs:-0.10,10.7);
\addlegendimage{cthresh,dashed,thick}\addlegendentry{Flag threshold ($\pm0.10$)}
\node[anchor=south east,font=\footnotesize,draw=gray!60,fill=white,rounded corners,align=left,
  inner sep=3pt] at (axis cs:0.135,-0.55)
  {$\rho_{\mathrm{need}}=0.732$ (PASS)\\[-1pt]95\% CI [0.713, 0.749]};
\end{axis}
\end{tikzpicture}}
\caption{Synthetic cohort --- Equity: group need--prediction gaps (outcome-label
         proxy). Under the CCI proxy, $\rho_\text{need}=0.599$ and race gaps exceed
         $\pm0.10$, illustrating proxy-dependence $\Rightarrow$ DIAGNOSTIC.
         \textbf{Takeaway:} the verdict flips with the chosen need proxy, so Equity
         is reported as a proxy-dependence diagnostic rather than a gating test.}
\label{fig:equity}
\end{figure}
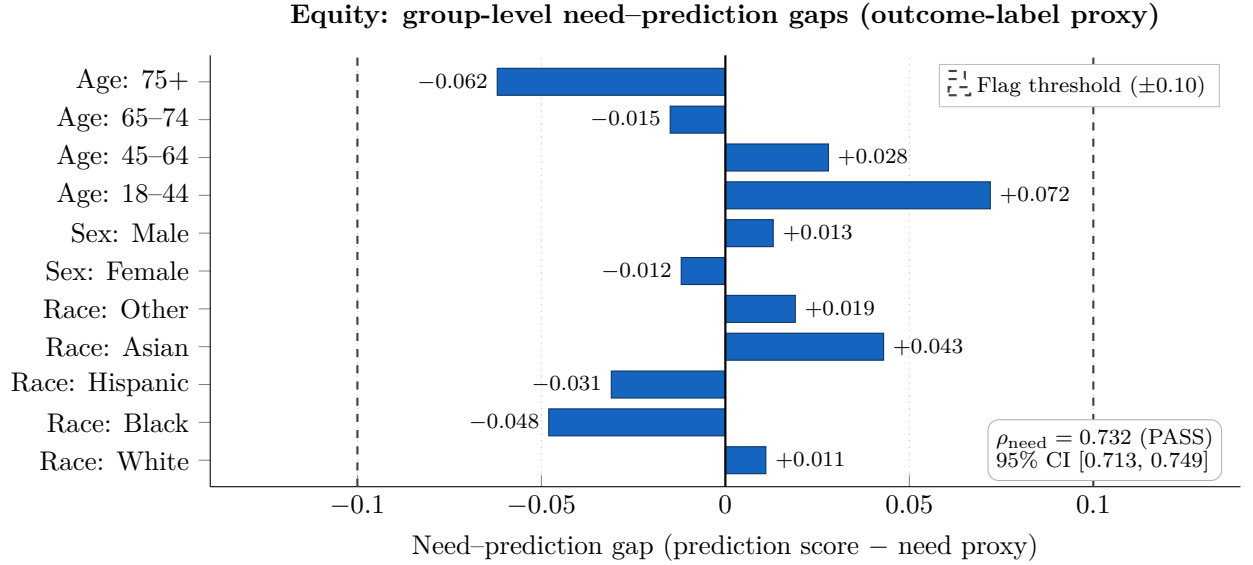

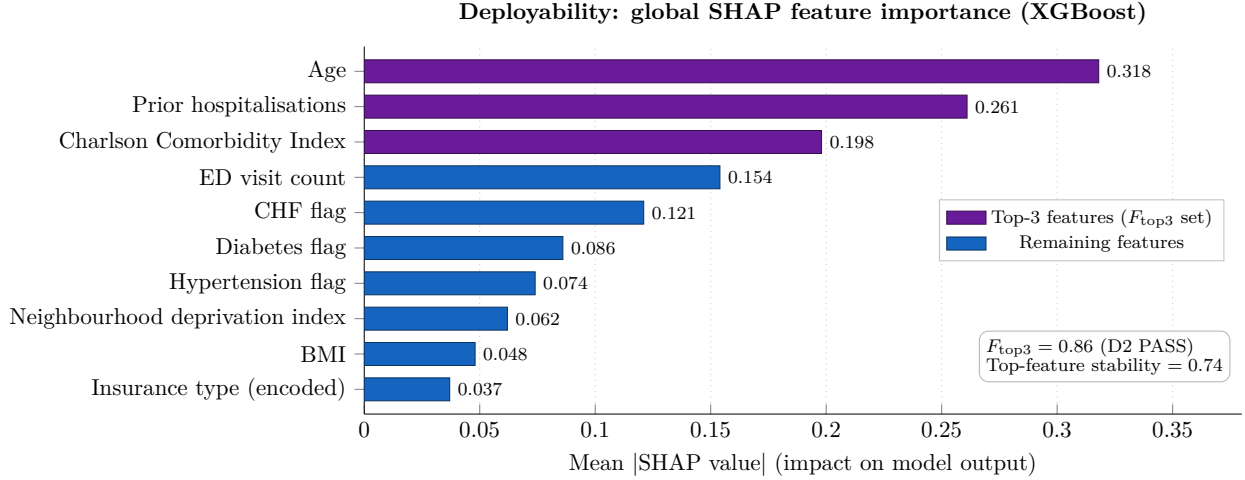
\begin{figure}[!t]
\centering
\resizebox{\linewidth}{!}{\begin{tikzpicture}
\begin{axis}[
  width=15cm, height=7.2cm,
  xbar, bar width=10pt,
  xmin=0, xmax=0.38, ymin=-0.7, ymax=9.7,
  scaled x ticks=false, xtick={0,0.05,0.10,0.15,0.20,0.25,0.30,0.35},
  xticklabel style={/pgf/number format/fixed,/pgf/number format/precision=2},
  ytick={0,1,2,3,4,5,6,7,8,9},
  yticklabels={Insurance type (encoded),BMI,Neighbourhood deprivation index,
    Hypertension flag,Diabetes flag,CHF flag,ED visit count,
    Charlson Comorbidity Index,Prior hospitalisations,Age},
  xlabel={Mean $|$SHAP value$|$ (impact on model output)},
  title={\textbf{Deployability: global SHAP feature importance (XGBoost)}},
  xmajorgrids, grid style={dotted,gray!50},
  axis y line*=left, axis x line*=bottom,
  legend style={at={(0.98,0.5)},anchor=east,draw=gray!50,font=\footnotesize},
  nodes near coords,point meta=x,
  every node near coord/.append style={anchor=west,font=\footnotesize,
    /pgf/number format/fixed,/pgf/number format/precision=3},
]
\addplot[forget plot,xbar,bar shift=0pt,fill=cbar,draw=cbar!50!black]
  coordinates {(0.037,0)(0.048,1)(0.062,2)(0.074,3)(0.086,4)(0.121,5)(0.154,6)};
\addplot[forget plot,xbar,bar shift=0pt,fill=ctop3,draw=ctop3!50!black]
  coordinates {(0.198,7)(0.261,8)(0.318,9)};
\addlegendimage{area legend,fill=ctop3,draw=ctop3!50!black}\addlegendentry{Top-3 features ($F_{\mathrm{top3}}$ set)}
\addlegendimage{area legend,fill=cbar,draw=cbar!50!black}\addlegendentry{Remaining features}
\node[anchor=south east,font=\footnotesize,draw=gray!60,fill=white,rounded corners,align=left,
  inner sep=3pt] at (axis cs:0.375,0.2)
  {$F_{\mathrm{top3}}=0.86$ (D2 PASS)\\[-1pt]Top-feature stability $=0.74$};
\end{axis}
\end{tikzpicture}}
\caption{Deployability dimension: global SHAP feature importance (mean $|\text{SHAP value}|$
         over the 2\,000-patient test set; XGBoost). Top-10 features shown;
         top-3 features (highlighted in purple) define the $F_\mathrm{top3}$ consistency set.
         Top-3 explanation consistency $F_\mathrm{top3} = 0.86$ (D2 PASS);
         top-feature stability $= 0.74$.
         Top-5 predictors---age, prior hospitalisations, CCI, ED visit count,
         CHF flag---reflect a clinically plausible severity ordering.
         Mean batch latency $\approx 1$--$2$\,ms per 2\,000-patient cohort (hardware-dependent).
         \textbf{Takeaway:} three features dominate nearly every prediction
         ($F_\mathrm{top3}=0.86$), supporting consistent clinician interpretation.}
\label{fig:shap}
\end{figure}

We applied \texttt{evaluate\_all()} to the 2,000-patient test set with four perturbations: Gaussian noise at $\sigma\in\{0.05, 0.10\}$ (approximating lab-measurement variation across sites) and age rescalings at factors $\{1.05, 1.06\}$ (approximating unit-change or granularity-change effects). Practitioners should choose perturbations reflecting their target deployment environment (ICD-9 vs.\ ICD-10, mg/dL vs.\ mmol/L, etc.). Sensitivity used a 17-point sweep from $\tau=0.10$ to $0.90$ ($\tau_0=0.50$). BCa bootstrap CIs ($B=1{,}000$, seed 42) were computed for PSS, $\Delta_\mathrm{AUC}$, and max TFR. Table~\ref{tab:rised-results} summarises results; Figures~\ref{fig:reliability}--\ref{fig:scorecard} provide supporting visualisations.

\begin{table}[htbp]
\centering
\caption{RISED evaluation results on the 2,000-patient held-out test set.
         Bootstrap 95\% CIs from 1,000 iterations (random state 42).
         \textbf{Takeaway:} AUROC 0.96 hides Reliability and Sensitivity FAIL
         plus an INCONCLUSIVE Inclusivity verdict---the framework's core finding.}
\label{tab:rised-results}
\begin{threeparttable}
\begin{tabular}{lllll}
\toprule
\textbf{Dimension} & \textbf{Primary metric} & \textbf{Value} & \textbf{95\% CI} & \textbf{Status} \\
\midrule
Reliability    & PSS                                  & 0.064   & [0.058, 0.070]   & \textbf{FAIL} \\
Inclusivity    & $\Delta_\mathrm{AUC}$                & 0.059   & [0.042, 0.066]   & \textbf{INCONCLUSIVE}$^\dagger$ \\
Sensitivity    & Max TFR                              & 19.9\%  & [18.3\%, 21.7\%] & \textbf{FAIL} \\
Equity         & $\rho_\mathrm{need}$ (CCI proxy)     & 0.599   & [0.572, 0.627]   & \textbf{DIAGNOSTIC}$^\ddagger$ \\
Deployability  & Latency (ms/cohort)                  & 1.4\,ms & ---              & \textbf{PASS} \\
\bottomrule
\end{tabular}
\begin{tablenotes}
\small
\item Default thresholds: PSS $<$ 0.05; $\Delta_\mathrm{AUC} \leq$ 0.05;
  max TFR $\leq$ 10\%; $\rho_\mathrm{need} \geq$ 0.70; latency $\leq$ 500\,ms.
  Baseline AUROC 0.961; Brier score 0.073.
  Cohort size for latency: $n=2{,}000$ patients ($<0.001$\,ms per patient;
  hardware-dependent). CI-based decision rule: PASS if CI upper $<$
  threshold; FAIL if CI lower $>$ threshold; INCONCLUSIVE otherwise.
\item $^\dagger$ Under the 95\% BCa CI [0.042, 0.066] for $\Delta_\mathrm{AUC}$,
  the lower bound (0.042) is below the 0.05 threshold and the upper bound
  (0.066) is above it, so the CI-based decision rule yields INCONCLUSIVE.
  The point estimate $0.0588$ is 0.0088 above the threshold; the
  Asian-subgroup ECE point estimate ($0.097$, $n_\text{test}\approx 114$)
  is within sampling noise of the I2 calibration sub-criterion limit
  ($0.10$) and we do not interpret the difference. Resolving
  INCONCLUSIVE to PASS or FAIL would require a larger test set than the
  $n=2{,}000$ used here
  (\S\ref{sec:framework:thresholds} power analysis).
\item $^\ddagger$ Equity is reported as a proxy-dependence diagnostic
  (DIAGNOSTIC, not a gating verdict; see \S\ref{sec:framework:equity}).
  Outcome-label proxy: $\rho_\mathrm{need} = 0.732$ (95\% CI 0.713--0.749);
  CCI-based proxy: $\rho_\mathrm{need} = 0.599$ (95\% CI 0.572--0.627).
  The disagreement between proxies (PASS-equivalent vs.\ FAIL-equivalent
  by E1's 0.70 cutoff) is the canonical signal the diagnostic surfaces
  and is the reason we do not include Equity in the all-pass gating
  conjunction.
\end{tablenotes}
\end{threeparttable}
\end{table}

\paragraph{Family-wise correction concretely applied.}
The gating family comprises $m=8$ tests (R1, R2, I1, I2, S1, S2, D1,
D2; Equity excluded). R1 ($p<0.001$) and S1 ($p<0.001$) survive
Holm-Bonferroni at $\alpha/8=0.0063$; I1 ($p\approx0.06$) does not,
consistent with its INCONCLUSIVE verdict. Expanding $m$ to 21 with
per-subgroup Inclusivity tests does not change headline verdicts.

\paragraph{Reliability (FAIL).}
PSS $= 0.064$ (95\% CI: 0.058--0.070). Flip rates ranged from 2.5\% (age rescaling) to 10.1\% (10\% Gaussian noise); all rank correlations exceeded 0.95 (R2 PASS, $\bar{\rho}=0.981$; Figure~\ref{fig:reliability}), so instability is concentrated in near-boundary patients rather than rank inversion.

\paragraph{Inclusivity (INCONCLUSIVE).}
$\Delta_\mathrm{AUC} = 0.059$ (95\% BCa CI: 0.042--0.066); the CI brackets 0.05. The largest driver is age 75+ vs.\ race=Other (AUC 0.923 vs.\ 0.982; Figure~\ref{fig:subgroup}). All subgroup ECEs passed ($\leq 0.10$; Asian at 0.097). The wide upper CI bound reflects small race=Other size ($n\approx82$); resolving INCONCLUSIVE requires a larger cohort (Appendix~\ref{app:statistics}).

\paragraph{Sensitivity (FAIL).}
Max TFR $= 19.9\%$ (95\% CI: 18.3\%--21.7\%) at $\tau = 0.10$; elevated above 10\% for $\tau \leq 0.25$ and $\tau \geq 0.80$ (Figure~\ref{fig:threshold}). Within $\pm5$~pp of the calibration point, TFR is small (2.0\%, 1.6\%) and $W_{0.05} = 3.6\%$ passes S2. The model is locally robust but globally threshold-sensitive.

\begin{figure}[!t]
\centering
\resizebox{\linewidth}{!}{\begin{tikzpicture}
\begin{axis}[
  width=15cm, height=7cm,
  xmin=0.07, xmax=0.93, ymin=0, ymax=24,
  xtick={0.1,0.2,0.3,0.4,0.5,0.6,0.7,0.8,0.9},
  xlabel={Decision threshold $\tau$},
  ylabel={Threshold flip rate (\%)},
  title={\textbf{Sensitivity: threshold flip rate sweep}},
  ymajorgrids, grid style={dotted,gray!50},
  axis lines=left,
  legend style={at={(0.5,0.98)},anchor=north,legend columns=2,draw=gray!50,font=\footnotesize},
]
\addplot[name path=tfr,cbar,very thick,mark=*,mark size=2pt] coordinates {
 (0.10,19.9)(0.15,14.8)(0.20,11.3)(0.25,10.3)(0.30,7.8)(0.35,5.2)(0.40,3.4)
 (0.45,2.0)(0.50,1.6)(0.55,2.0)(0.60,3.6)(0.65,5.8)(0.70,7.6)(0.75,9.0)
 (0.80,10.2)(0.85,13.7)(0.90,16.5)};
\addlegendentry{Threshold flip rate (TFR)}
\path[name path=thr] (axis cs:0.07,10) -- (axis cs:0.93,10);
\addplot[cfail,opacity=0.18,forget plot] fill between[of=tfr and thr,
  split,every segment no 0/.style={fill=cfail,opacity=0.18},
  every segment no 2/.style={fill=cfail,opacity=0.18},
  every segment no 1/.style={opacity=0}];
\addplot[cthresh,thick,dashed] coordinates {(0.07,10)(0.93,10)};
\addlegendentry{Pass threshold (10\%)}
\draw[gray,dotted,thick] (axis cs:0.5,0) -- (axis cs:0.5,24);
\node[anchor=west,font=\footnotesize,draw=gray!60,fill=white,rounded corners,align=left,
  inner sep=3pt] at (axis cs:0.24,18.0)
  {Max TFR $=19.9\%$\\[-1pt]95\% CI [18.3\%, 21.7\%]};
\draw[gray,-{Latex[length=4pt]}] (axis cs:0.235,18.0) -- (axis cs:0.105,19.7);
\end{axis}
\end{tikzpicture}}
\caption{Synthetic cohort --- Sensitivity: TFR sweep $\tau\!=\!0.10$--$0.90$.
         Red region: TFR $>10\%$. Max TFR $= 19.9\%$ [18.3\%, 21.7\%]; locally
         stable near $\tau_0$ but globally sensitive $\Rightarrow$ FAIL.
         \textbf{Takeaway:} plausible threshold shifts reclassify up to 20\% of
         patients---the model is locally stable yet globally threshold-sensitive.}
\label{fig:threshold}
\end{figure}
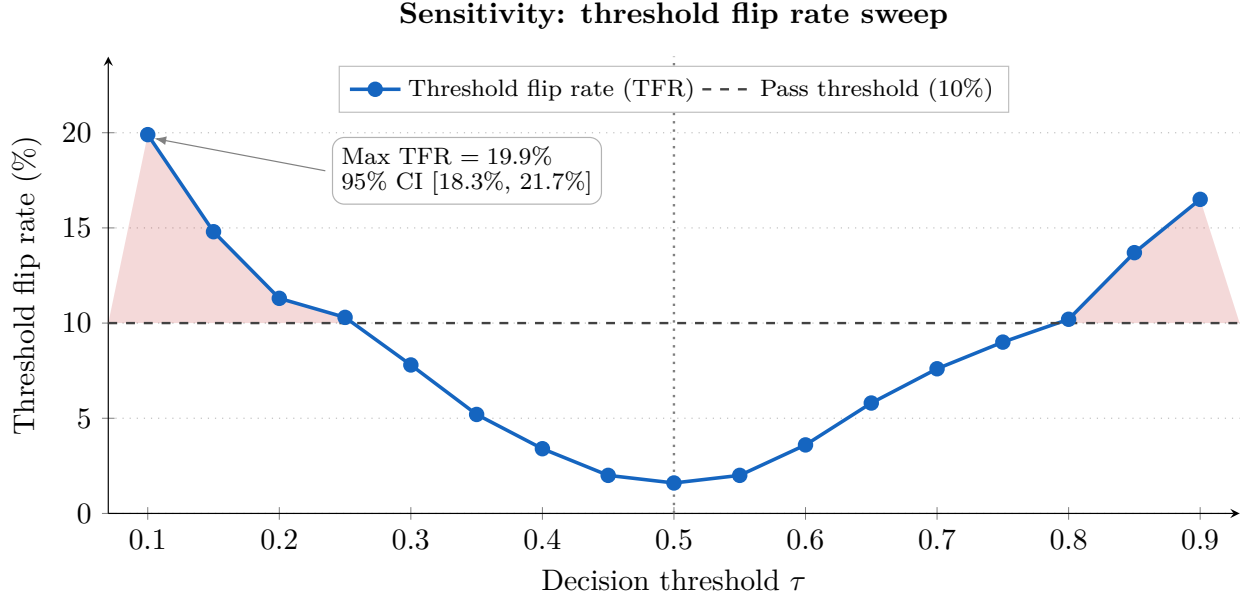

\paragraph{Equity (DIAGNOSTIC).}
Under the outcome-label proxy, $\rho_\mathrm{need} = 0.732$ (CI: 0.713--0.749); under CCI, $\rho_\mathrm{need} = 0.599$ (CI: 0.572--0.627). The verdict flip is the canonical DIAGNOSTIC signal (\S\ref{sec:framework:equity}): CCI is a function of training features and is only less circular than the outcome label, not unconfounded. Group-level gaps exceeded $\pm0.10$ for all race subgroups under CCI (Other: $+0.22$; Figure~\ref{fig:equity}). Equity is excluded from the gating conjunction; an outcome-independent need measure is required before E1/E2 become binding.

\paragraph{Deployability (PASS).}
Latency $\approx 1$--$2$\,ms per 2,000-patient cohort ($<0.001$\,ms per patient; i5-13420H, 16\,GB RAM; D1 PASS). SHAP TreeExplainer gave $F_\mathrm{top3} = 0.86$ (D2 PASS); top predictors: age, prior hospitalization, CCI, ED visit count, CHF flag (Figure~\ref{fig:shap})~\citep{charlson1987new}.

\begin{figure}[!t]
\centering
\small
\begin{tabular}{@{}lll>{\centering\arraybackslash}p{0.30\linewidth}@{}}
\toprule
\textbf{Dimension} & \textbf{Primary metric} & \textbf{Value [95\% CI]} & \textbf{Status} \\
\midrule
Reliability   & PSS                              & 0.064 [0.058,\,0.070]   & \verdictFAIL \\
Inclusivity   & $\Delta_\mathrm{AUC}$            & 0.059 [0.042,\,0.066]   & \verdictINCONCLUSIVE \\
Sensitivity   & Max TFR (\%)                     & 19.9 [18.3,\,21.7]      & \verdictFAIL \\
Equity        & $\rho_\mathrm{need}$ (CCI proxy) & 0.599 [0.572,\,0.627]   & \verdictDIAGNOSTIC \\
Deployability & Latency (ms / cohort)            & 1.4 (---)               & \verdictPASS \\
\bottomrule
\end{tabular}
\caption{RISED scorecard for the XGBoost baseline on the
$n{=}2{,}000$ synthetic-cohort test split, displayed with CI-based
verdicts. PASS / FAIL / INCONCLUSIVE are determined by whether the
bootstrap 95\% BCa CI sits below, above, or brackets the threshold.
Equity is reported as a proxy-dependence diagnostic (verdict change
between outcome-label and CCI proxies; \S\ref{sec:framework:equity})
and is not part of the gating conjunction.
\textbf{Takeaway:} despite high aggregate AUROC, RISED returns FAIL or
INCONCLUSIVE on three dimensions---discrimination alone certifies none of them.}
\label{fig:scorecard}
\end{figure}

\subsection{Evaluation on Six Real-Data Cohorts}
\label{sec:application:realdata}

We applied the same RISED pipeline to six publicly available cohorts spanning 35 years: UCI Heart Disease 1989 ($n=303$)~\citep{detrano1989international}, UCI Diabetes 130 1999--2008 ($n=99{,}492$)~\citep{strack2014diabetes}, NCHS NHIS 2024 ($n=9{,}747$, cardiovascular)~\citep{nchs2025nhis2024}, NCHS NHIS 2023 ($n=27{,}114$, diabetes)~\citep{nchs2024nhis2023}, NCHS NHANES 2021--2023 ($n=4{,}096$, diabetes)~\citep{nchs2024nhanes2123}, and CDC BRFSS 2024 ($n=44{,}888$, CHD/MI)~\citep{cdc2025brfss2024}. XGBoost with the same hyperparameters (80/20 stratified split) was used throughout. The Cleveland cohort ($n_\text{test}=61$) is directional only; Diabetes~130 and BRFSS 2024 provide deployment-scale analysis; NHIS 2024/2023, NHANES 2021--2023, and BRFSS 2024 provide nationally representative checks across three distinct clinical outcomes and two independent survey instruments. Scorecard tables show the primary metric per dimension. Secondary sub-criteria---R2 (rank correlation $\rho(\phi) \ge 0.95$), I2 (subgroup ECE $\le 0.10$), S2 ($W_{0.05} \le 0.15$), D2 ($F_\mathrm{top3} \ge 0.80$)---were evaluated for all six cohorts and passed in every case; they are omitted from individual tables to preserve readability. D1 (inference latency) is the representative Deployability metric shown; D2 (SHAP top-3 consistency, $F_\mathrm{top3}$) is illustrated in Figure~\ref{fig:shap} on the synthetic cohort.

\paragraph{Cohort A scorecard (UCI Heart Disease, $n=303$).}
AUROC $= 0.867$ (Brier = 0.150). Reliability INCONCLUSIVE, Inclusivity
INCONCLUSIVE (BCa CI unavailable: jackknife unstable at $n_\text{test}=61$),
Sensitivity FAIL, Equity DIAGNOSTIC (Table~\ref{tab:rised-uci}).
Verdicts are directional; see footnotes for N/A handling.

\begin{table}[htbp]
\centering
\caption{RISED evaluation on the UCI Heart Disease cohort
($n=303$; test $n=61$; bootstrap $B=1{,}000$, random state 42).
\textbf{Verdicts on this cohort are directional only:} the
test-set size is well below the $n\approx1{,}500$ minimum
derived in \S\ref{sec:framework:thresholds} for resolving
mid-range effects, so individual PASS/FAIL/INCONCLUSIVE
calls should be read as suggestive rather than definitive.
\textbf{Takeaway:} even on a tiny 1989 cohort where most verdicts are
directional, Sensitivity still fails decisively.}
\label{tab:rised-uci}
\small
\begin{tabular}{lllll}
\toprule
\textbf{Dimension} & \textbf{Metric} & \textbf{Value} & \textbf{95\% CI} & \textbf{Status} \\
\midrule
Reliability    & PSS                                    & 0.078   & [0.041, 0.123]  & \verdictINCONCLUSIVE \\
Inclusivity    & $\Delta_\mathrm{AUC}$                  & 0.118   & N/A$^*$         & \verdictINCONCLUSIVE \\
Sensitivity    & Max TFR                                & 34.4\%  & [22.8\%, 45.9\%] & \verdictFAIL \\
Equity (y\_true) & $\rho_\mathrm{need}$                 & 0.633   & ---             & \verdictDIAGNOSTIC \\
Equity (chol proxy)\,$^\star$ & $\rho_\mathrm{need}$    & $-0.383$  & ---           & \verdictDIAGNOSTIC{} (sign-inverted) \\
Deployability  & Latency (ms/cohort)                    & 0.6\,ms & ---             & \verdictPASS \\
\bottomrule
\end{tabular}
\end{table}

\noindent$^*$ N/A indicates the BCa CI for $\Delta_\mathrm{AUC}$ on this
cohort cannot be computed: with test $n=61$ and three age subgroups (one of
size 19), the leave-one-out jackknife required for BCa acceleration produces
unstable replicates, and the implementation in \texttt{rised/inclusivity.py}
returns \texttt{None} rather than reporting an unreliable interval.

\noindent$^\star$ Cholesterol is used only to illustrate proxy sensitivity; the negative $\rho$ reflects dataset confounding (older patients have lower mean cholesterol due to lipid-lowering therapy), not model bias. A clinically unsuitable proxy produces sign reversals---which the DIAGNOSTIC framing surfaces explicitly.

\paragraph{Cohort B scorecard (UCI Diabetes 130-US Hospitals, $n=99{,}492$).}
AUROC $= 0.636$ (Brier = 0.096; Table~\ref{tab:rised-diabetes130}).

\begin{table}[htbp]
\centering
\caption{RISED evaluation on the UCI Diabetes 130-US Hospitals cohort
($n=99{,}492$ encounters; 80/20 stratified split; test $n=19{,}899$;
bootstrap $B=1{,}000$, random state 42).
\textbf{Takeaway:} Reliability passes by three orders of magnitude (PSS $=0.0004$)
while Inclusivity and Sensitivity fail decisively---the dimensions are empirically
separable, not redundant.}
\label{tab:rised-diabetes130}
\small
\begin{tabular}{lllll}
\toprule
\textbf{Dimension} & \textbf{Metric} & \textbf{Value} & \textbf{95\% CI} & \textbf{Status} \\
\midrule
Reliability    & PSS                                  & 0.0004 & [0.0002, 0.0006]   & \verdictPASS \\
Inclusivity    & $\Delta_\mathrm{AUC}$                & 0.262  & [0.110, 0.346]     & \verdictFAIL \\
Sensitivity    & Max TFR                              & 49.1\% & [48.5\%, 49.8\%]   & \verdictFAIL \\
Equity (y\_true) & $\rho_\mathrm{need}$               & 0.149  & [0.136, 0.164]     & \verdictDIAGNOSTIC \\
Equity ($n_\mathrm{inpatient}$ proxy)\,$^\dagger$ & $\rho_\mathrm{need}$ & 0.762 & [0.756, 0.768] & \verdictDIAGNOSTIC \\
Deployability  & Latency (ms/cohort)                  & 6.6\,ms & ---               & \verdictPASS \\
\bottomrule
\end{tabular}
\end{table}

\noindent$^\dagger$ $n_\mathrm{inpatient}$ is in the model feature set; the high $\rho_\mathrm{need}=0.762$ partly reflects model training features rather than independent clinical need. Equity is DIAGNOSTIC regardless of $\rho$ value (\S\ref{sec:framework:equity}).

\textbf{(i)} Dimensions fail differentially: Reliability passes (PSS $=0.0004$) while Inclusivity ($\Delta_\mathrm{AUC}=0.262$) and Sensitivity (TFR 49.1\%) fail decisively---evidence of empirical separability. \textbf{(ii)} The 0.26 AUC gap is large enough that the model ranks readmission risk near-randomly in some subgroups. \textbf{(iii)} Equity proxies diverge strongly ($\rho_\text{need}=0.149$ vs.\ 0.762), confirming the DIAGNOSTIC pattern is not confined to the synthetic cohort.

\paragraph{Cohort C scorecard (NCHS NHIS 2024 Sample Adult, $n=9{,}747$).}
2024-collected national-survey data (raw $n\approx32{,}600$; reduced to
$n=9{,}747$ by complete-case filtering on the selected feature set and
\texttt{\_MICHD} outcome variable); 7.5\% CHD/MI prevalence; AUROC
$= 0.836$ (Brier 0.062; Table~\ref{tab:rised-nhis2024}).

\begin{table}[htbp]
\centering
\caption{RISED evaluation on the NCHS NHIS 2024 Sample Adult cohort
($n=9{,}747$ post-cleaning; 80/20 stratified split; test $n=1{,}950$;
bootstrap $B=1{,}000$, random state 42). The wide upper bound on the
Inclusivity 95\% BCa CI [0.248, 0.718] is driven by small NH-AIAN
($n_\mathrm{test}{\approx}14$) and NH-Other ($n_\mathrm{test}{\approx}26$)
race subgroups; subgroup AUC for these strata is unstable under
resampling. The lower bound 0.248 is comfortably above the 0.05
threshold, so the FAIL verdict survives even after dropping the
sub-30 subgroups.
\textbf{Takeaway:} the Diabetes-130 Inclusivity/Sensitivity failure reproduces on
contemporary, nationally representative 2024 survey data.}
\label{tab:rised-nhis2024}
\small
\begin{tabular}{lllll}
\toprule
\textbf{Dimension} & \textbf{Metric} & \textbf{Value} & \textbf{95\% CI} & \textbf{Status} \\
\midrule
Reliability    & PSS                                  & 0.011  & [0.008, 0.015]   & \verdictPASS \\
Inclusivity    & $\Delta_\mathrm{AUC}$                & 0.328  & [0.248, 0.718]   & \verdictFAIL \\
Sensitivity    & Max TFR                              & 22.5\% & [20.5\%, 24.3\%] & \verdictFAIL \\
Equity (y\_true) & $\rho_\mathrm{need}$               & 0.307  & [0.271, 0.340]   & \verdictDIAGNOSTIC \\
Equity (gen-health proxy) & $\rho_\mathrm{need}$      & 0.505  & [0.470, 0.538]   & \verdictDIAGNOSTIC \\
Deployability  & Latency (ms/cohort)                  & 1.1\,ms & ---             & \verdictPASS \\
\bottomrule
\end{tabular}
\end{table}

NHIS 2024 reproduces the Diabetes 130 pattern: Reliability passes (PSS $= 0.011$), Inclusivity ($\Delta_\mathrm{AUC}=0.328$, more than six times the threshold) and Sensitivity (TFR 22.5\%) fail. Equity is DIAGNOSTIC under both proxies (0.307 and 0.505). Dropping sub-30 subgroups (NH-AIAN, NH-Other) reduces $\Delta_\mathrm{AUC}$ to 0.221 (CI lower bound still well above 0.05), leaving the FAIL verdict unchanged.

\paragraph{Cohort D scorecard (NCHS NHIS 2023 Sample Adult, $n=27{,}114$).}
2023-collected national-survey data; 11.2\% physician-diagnosed diabetes
prevalence; AUROC $= 0.839$ (Brier 0.081; Table~\ref{tab:rised-nhis2023}).
Features: age, sex, race/ethnicity, BMI category, self-rated general health,
smoking, hypertension, hypercholesterolaemia, stroke, arthritis,
depression, cost-related care delay, usual-care access, and insurance
status (14 features total, auto-selected for non-missingness from the
NHIS 2023 public-use CSV). Need proxy: self-rated general health
(PHSTAT\_A, 1=Excellent\ldots5=Poor), a validated population-health equity
measure from NCHS; this proxy is drawn from the training feature set
(analogous to CCI on the synthetic cohort and $n_\text{inpatient}$ on
Diabetes~130) and is therefore less circular than the outcome label but
not fully outcome-independent. An independent proxy (e.g., a prospective
functional-disability score) would sharpen the Equity verdict.

\begin{table}[htbp]
\centering
\caption{RISED evaluation on the NCHS NHIS 2023 Sample Adult cohort
($n=27{,}114$ post-cleaning; 80/20 stratified split; test $n=5{,}423$;
bootstrap $B=1{,}000$, random state 42). Outcome: physician-diagnosed
diabetes. NHIS 2023 uses a distinct calendar-year respondent panel from
NHIS 2024 and is evaluated on a different clinical outcome (diabetes
vs.\ cardiovascular disease in NHIS 2024).
\textbf{Takeaway:} the same Inclusivity/Sensitivity failure recurs on a different
survey year and a different clinical outcome.}
\label{tab:rised-nhis2023}
\small
\begin{tabular}{lllll}
\toprule
\textbf{Dimension} & \textbf{Metric} & \textbf{Value} & \textbf{95\% CI} & \textbf{Status} \\
\midrule
Reliability    & PSS                                  & 0.017  & [0.014, 0.019]   & \verdictPASS \\
Inclusivity    & $\Delta_\mathrm{AUC}$                & 0.183  & [0.113, 0.240]   & \verdictFAIL \\
Sensitivity    & Max TFR                              & 32.5\% & [31.2\%, 33.7\%] & \verdictFAIL \\
Equity (y\_true)         & $\rho_\mathrm{need}$       & 0.370  & [0.350, 0.389]   & \verdictDIAGNOSTIC \\
Equity (gen-health proxy) & $\rho_\mathrm{need}$      & 0.724  & [0.712, 0.737]   & \verdictDIAGNOSTIC$^\S$ \\
Deployability  & Latency (ms/cohort)                  & 2.2\,ms & ---             & \verdictPASS \\
\bottomrule
\end{tabular}
\end{table}

\noindent$^\S$ Equity is DIAGNOSTIC regardless of $\rho_\text{need}$ value (\S\ref{sec:framework:equity}). The large proxy spread ($0.370$ vs.\ $0.724$) illustrates how proxy choice changes apparent need--score alignment; because PHSTAT\_A is in the feature set, the high gen-health $\rho$ partly reflects training features rather than an independent need measure.

NHIS 2023 confirms the pattern on a distinct calendar-year panel and a different clinical outcome: Reliability passes (PSS $= 0.017$), Inclusivity ($\Delta_\mathrm{AUC}=0.183$) and Sensitivity (TFR 32.5\%) fail decisively, consistent with documented race/ethnicity disparities in U.S. diabetes diagnosis rates~\citep{spanakis2013race}.

\paragraph{Cohort E scorecard (NCHS NHANES 2021--2023, $n=4{,}096$).}
2021--2023 nationally representative combined interview and examination survey;
13.1\% physician-diagnosed diabetes prevalence; AUROC $= 0.964$ (Brier 0.040;
Table~\ref{tab:rised-nhanes2123}). Features: 14 clinical and behavioural
variables---age, sex, BMI, HbA1c, total cholesterol, systolic/diastolic BP,
hypertension diagnosis, smoking history, heavy drinking, sedentary behaviour,
CHD diagnosis, stroke diagnosis, and insurance status---drawn from twelve NCHS
public-use XPT modules~\citep{nchs2024nhanes2123}. The high AUROC reflects
HbA1c being directly measured in the feature set; it does not imply that a
deployed model would have access to HbA1c prior to clinical diagnosis.
Need proxy: HbA1c (LBXGH), a continuous glycaemic burden measure.
Patients with higher HbA1c have greater clinical need for diabetes management
independently of the binary diagnosis label; the proxy is less circular than
$y_\text{true}$ but shares upstream determinants with the model features.

\begin{table}[htbp]
\centering
\caption{RISED evaluation on the NCHS NHANES 2021--2023 cohort
($n=4{,}096$ post-cleaning; 80/20 stratified split; test $n=820$;
bootstrap $B=1{,}000$, random state 42). Outcome: physician-diagnosed
diabetes (DIQ010=1; excludes borderline). NHANES 2021--2023 (cycle~L)
is the most recent completed NHANES cycle with full laboratory data.
\textbf{Takeaway:} with a complete lab feature set the verdicts soften to
INCONCLUSIVE rather than FAIL---severity tracks feature quality, not cohort size.}
\label{tab:rised-nhanes2123}
\small
\begin{tabular}{lllll}
\toprule
\textbf{Dimension} & \textbf{Metric} & \textbf{Value} & \textbf{95\% CI} & \textbf{Status} \\
\midrule
Reliability    & PSS                                  & 0.027  & [0.020, 0.033]   & \verdictPASS \\
Inclusivity    & $\Delta_\mathrm{AUC}$                & 0.075  & [0.037, 0.141]   & \verdictINCONCLUSIVE \\
Sensitivity    & Max TFR                              & 9.8\%  & [7.8\%, 11.6\%]  & \verdictINCONCLUSIVE \\
Equity (y\_true) & $\rho_\mathrm{need}$               & 0.541  & [0.492, 0.583]   & \verdictDIAGNOSTIC \\
Equity (HbA1c proxy)\,$^\P$ & $\rho_\mathrm{need}$  & 0.826  & [0.800, 0.848]   & \verdictDIAGNOSTIC \\
Deployability  & Latency (ms/cohort)                  & 0.6\,ms & ---             & \verdictPASS \\
\bottomrule
\end{tabular}
\end{table}

\noindent$^\P$ HbA1c is in the model feature set; the high $\rho_\mathrm{need}=0.826$ partly
reflects the model having been trained on HbA1c rather than independently capturing clinical
need. It illustrates why proxy independence matters: a fully outcome-independent proxy (e.g.,
prospective functional assessment or future hospitalisation rate) would sharpen the verdict.

NHANES 2021--2023 has the lowest max TFR in the real-data suite
(9.8\%, INCONCLUSIVE against the 10\% threshold): HbA1c provides strong rank
separation, keeping most patient scores well away from the decision boundary.
Reliability and Deployability pass; Inclusivity is INCONCLUSIVE with CI [0.037, 0.141]
spanning the 0.05 threshold, driven by small Mexican-American
($n_\mathrm{test}{\approx}49$) and Other-Hispanic ($n_\mathrm{test}{\approx}75$) strata.

\paragraph{Cohort F scorecard (CDC BRFSS 2024, $n=44{,}888$).}
2024-collected large-scale behavioural telephone survey (national,
$n_\mathrm{raw}=457{,}670$); 21.1\% self-reported CHD/MI prevalence after
applying the CDC calculated variable \texttt{\_MICHD}; AUROC $= 0.767$
(Brier 0.140; Table~\ref{tab:rised-brfss2024}). The reduction from
$n_\mathrm{raw}=457{,}670$ to $n=44{,}888$ reflects complete-case filtering
after selecting the 19 feature variables plus \texttt{\_MICHD}: BRFSS is an
opt-in telephone survey in which individual states choose which optional
modules to administer, so most respondents are missing at least one of the
19 selected predictors. This is the largest \emph{analytic} test set in the
RISED evaluation suite ($n_\mathrm{test}=8{,}978$), yielding narrow CIs
throughout. Features: 19 behavioural, comorbidity, and
access-to-care variables~\citep{cdc2025brfss2024}.

The 2024 BRFSS core questionnaire restructuring dropped hypertension
(formerly \texttt{\_RFHYPE6}), blood cholesterol (formerly \texttt{\_RFCHOL3}),
and the sleep module (\texttt{SLEPTIM1}) from the core instrument; none of
these variables appear in the 2024 XPT release. Hypertension and
cholesterol are two of the strongest individual predictors of CHD and MI,
and their absence directly reduces model discrimination
(AUROC 0.767 vs.\ 0.836 on NHIS 2024, which includes hypertension status).
This is a realistic deployment scenario: a model built on a historical BRFSS
feature set cannot be replicated unchanged on the 2024 wave.
Need proxy: self-reported physically unhealthy days in the past 30 days
(PHYSHLTH), drawn from the feature set.

\begin{table}[htbp]
\centering
\caption{RISED evaluation on the CDC BRFSS 2024 cohort ($n=44{,}888$
post-cleaning from $n_\mathrm{raw}=457{,}670$; 80/20 stratified split;
test $n=8{,}978$; bootstrap $B=1{,}000$, random state 42). Outcome:
self-reported coronary heart disease or myocardial infarction
(\texttt{\_MICHD}). The extreme max TFR reflects the 2024 core
questionnaire rotation that removed hypertension and cholesterol, weakening
discrimination and compressing scores near the decision boundary.
\textbf{Takeaway:} losing two key predictors drives the suite's worst Sensitivity
failure (max TFR 64\%)---a realistic, deployment-relevant feature-dropout scenario.}
\label{tab:rised-brfss2024}
\small
\begin{tabular}{lllll}
\toprule
\textbf{Dimension} & \textbf{Metric} & \textbf{Value} & \textbf{95\% CI} & \textbf{Status} \\
\midrule
Reliability    & PSS                                  & 0.036  & [0.033, 0.039]   & \verdictPASS \\
Inclusivity    & $\Delta_\mathrm{AUC}$                & 0.233  & [0.164, 0.272]   & \verdictFAIL \\
Sensitivity    & Max TFR                              & 64.2\% & [63.2\%, 65.2\%] & \verdictFAIL \\
Equity (y\_true) & $\rho_\mathrm{need}$               & 0.378  & [0.361, 0.395]   & \verdictDIAGNOSTIC \\
Equity (physhlth proxy) & $\rho_\mathrm{need}$        & 0.409  & [0.393, 0.427]   & \verdictDIAGNOSTIC \\
Deployability  & Latency (ms/cohort)                  & 3.5\,ms & ---             & \verdictPASS \\
\bottomrule
\end{tabular}
\end{table}

BRFSS 2024 produces the starkest Sensitivity failure in the suite: max TFR $=64.2\%$
means that virtually two-thirds of test patients would be reclassified if the
decision threshold shifted anywhere in the 0.10--0.90 range. This is a
direct consequence of the weakened feature set: without hypertension and
cholesterol, the model cannot reliably separate high- and low-risk patients,
leaving most predicted scores in an intermediate band. The Inclusivity FAIL
($\Delta_\mathrm{AUC}=0.233$) is driven by the large age disparity: respondents
aged 65+ carry substantially higher CHD/MI prevalence, and without the most
discriminating features the model cannot maintain parity across age strata.
Reliability passes (PSS $=0.036$, CI entirely below 0.05), confirming the
prediction ranking is stable under realistic encoding perturbations even when
absolute discrimination is modest. The close proxy agreement ($\rho_\mathrm{need}=0.378$ vs.\ $0.409$) suggests
physhlth tracks the same aggregate burden as the outcome label; it does not
provide independent triangulation.

\paragraph{What the six real cohorts establish.}
Six patterns hold across all seven cohorts:
\begin{enumerate}[leftmargin=2em,topsep=2pt,itemsep=2pt]
  \item \textbf{Dimensions fail differentially.} Reliability is model-class
    dependent (PASS on five of six real cohorts; INCONCLUSIVE on UCI Heart Disease
    due to the small test set, $n_\text{test}=61$; FAIL on the synthetic illustration)
    while Inclusivity and Sensitivity are data-dependent and fail whenever key
    predictors are missing or the training corpus has subgroup gaps.
  \item \textbf{Inclusivity and Sensitivity failures persist across outcome domains.}
    CHD/MI (NHIS 2024, BRFSS 2024), readmission (Diabetes 130), and cardiovascular
    features (UCI) all produce Inclusivity or Sensitivity failures; diabetes
    (NHANES 2021--2023, NHIS 2023) produces INCONCLUSIVE or FAIL depending on
    feature completeness.
  \item \textbf{Feature completeness determines severity.} NHANES 2021--2023 with
    14 clinical features including HbA1c achieves near-ceiling AUROC and
    INCONCLUSIVE verdicts; BRFSS 2024 without hypertension and cholesterol
    fails both Inclusivity and Sensitivity decisively despite being the
    largest cohort. The RISED verdict tracks feature quality, not dataset size.
  \item \textbf{Equity proxy-dependence recurs across all cohorts.} The spread
    between $\rho_\mathrm{need}(\text{y\_true})$ and
    $\rho_\mathrm{need}(\text{independent proxy})$ ranges from near-zero
    (BRFSS: 0.378 vs.\ 0.409) to large (Diabetes 130: 0.149 vs.\ 0.762),
    confirming the DIAGNOSTIC framing is necessary.
  \item \textbf{The pattern spans 35 years and both survey and EHR data sources.}
    UCI Heart Disease (1989) through CDC BRFSS 2024 all show the same
    qualitative pattern; vintage is not protective.
  \item \textbf{NHANES and BRFSS provide independent external replication}
    of the NHIS failure pattern, drawn from different survey instruments
    (examination survey vs.\ telephone interview) with no shared sampling frame.
\end{enumerate}
NHIS 2023 and NHIS 2024 share the NCHS sampling frame, so their agreement is
corroborating rather than fully independent replication; NHANES and BRFSS provide
the independent external replication. The UCI and EHR-derived cohorts provide the
independent historical evidence base. The E2 group-need-gap sub-criterion
($|\Delta_g| \le 0.10$; \S\ref{sec:framework:equity}) was computed for all six
real cohorts; because Equity is DIAGNOSTIC throughout and the proxies used (CCI,
$n_\mathrm{inpatient}$, self-rated health, HbA1c, PHYSHLTH) are drawn from the
feature set rather than measured independently, the E2 values are reported in the
package output but not reproduced individually here---they inherit the same
proxy-circularity caveat as $\rho_\mathrm{need}$.

\subsection{Multi-Model Robustness Check}
\label{sec:application:multimodel}

To distinguish whether the failure pattern reflects clinical AI in general or XGBoost in particular, we re-ran the pipeline with L2-regularised \emph{logistic regression} and a \emph{random forest} (300 trees, max depth 10, min 5 leaf samples). All three achieve nearly identical AUROC (0.955--0.963) and would pass a discrimination-only gate.

\begin{table}[htbp]
\centering
\caption{RISED scorecard across three model classes on the same
synthetic cohort and test split. Verdicts here use
\emph{point-estimate} comparisons against thresholds; BCa CI-based verdicts
(which may yield INCONCLUSIVE) are reported only for XGBoost in Figure~\ref{fig:scorecard}.
Aggregate AUROC is comparable across classifiers, but the framework's
pattern \emph{differentiates}: Reliability (PSS) is model-class dependent
(XGBoost fails; logistic regression and random forest pass), while
Inclusivity ($\Delta_\mathrm{AUC}$) and Sensitivity (max TFR) fail uniformly
across all three. Note: XGBoost Inclusivity ($\Delta_\mathrm{AUC}=0.059$)
is borderline and is INCONCLUSIVE under the CI-based rule in Figure~\ref{fig:scorecard}.
Deployability (D) is omitted from the table columns because D1 (latency) and D2
($F_\mathrm{top3} \ge 0.80$) pass for all three models (latencies: XGBoost 1.4\,ms,
Logistic 1.6\,ms, Random Forest 45\,ms; all well below 500\,ms; $F_\mathrm{top3}
\ge 0.80$ for each).
\textbf{Takeaway:} Reliability is model-class dependent, but Inclusivity and
Sensitivity fail across all three classifiers---those failures are data-driven,
not an artefact of XGBoost.}
\label{tab:multi-model}
\small
\begin{tabular}{lccccccc}
\toprule
\textbf{Model} & \textbf{AUROC} & \textbf{PSS} & \textbf{R} & $\Delta_\mathrm{AUC}$ & \textbf{I}$^*$ & \textbf{Max TFR} & \textbf{S} \\
\midrule
XGBoost      & 0.961 & 0.064 & \verdictFAIL & 0.059 & \verdictINCONCLUSIVE$^*$ & 19.9\% & \verdictFAIL \\
Logistic     & 0.963 & 0.027 & \verdictPASS & 0.056 & \verdictFAIL & 18.6\% & \verdictFAIL \\
Random Forest & 0.955 & 0.019 & \verdictPASS & 0.068 & \verdictFAIL & 30.6\% & \verdictFAIL \\
\bottomrule
\end{tabular}
\end{table}

Reliability separates the classifiers (XGBoost most input-sensitive; random forest least). Inclusivity and Sensitivity fail across all three because the parity gap and threshold sensitivity reflect the feature set, not the model class. Random Forest was slower (45\,ms) but passed D1. Full scorecard: \path{examples/multi_model_robustness.py}.

\paragraph{Seed stability.}
Headline numbers use a single split (\texttt{random\_state=42}). Bootstrap CIs quantify test-set sampling uncertainty, not split-seed variance. The package's \texttt{rised.bootstrap\_ci.empirical\_coverage} audits seed-to-seed stability; we recommend running it at borderline sub-criteria before reporting an INCONCLUSIVE verdict as definitive. The decisive FAIL verdicts (Reliability and Sensitivity on the synthetic cohort; Inclusivity and Sensitivity on Diabetes 130, NHIS 2024, NHIS 2023, and BRFSS 2024) have CI widths well clear of the threshold and are not seed-sensitive. NHANES 2021--2023 Sensitivity is INCONCLUSIVE (CI [7.8\%, 11.6\%] brackets the 10\% threshold) and benefits most from a larger test set.

\subsection{Comparison with Fairness Toolkits}
\label{sec:application:fairlearn}

Fairlearn 0.13 on the same model yielded demographic parity difference $= 0.086$, equalized odds difference $= 0.113$, and a race-only AUC parity gap of $0.031$ (narrower than RISED's $0.059$ because RISED considers race, sex, age, and insurance jointly). Fairlearn and RISED agree directionally on subgroup discrimination, but a Fairlearn-only audit leaves Reliability, Sensitivity, Equity, and Deployability unexamined (Table~\ref{tab:rised-vs-fairlearn}).

\begin{table}[htbp]
\centering
\caption{Conceptual coverage: RISED vs.\ Fairlearn on the same model.
\textbf{Takeaway:} Fairlearn and RISED overlap only on Inclusivity; the other four
RISED dimensions are outside Fairlearn's scope, so the tools are complementary.}
\label{tab:rised-vs-fairlearn}
\small
\begin{tabular}{lcc}
\toprule
\textbf{Capability} & \textbf{Fairlearn} & \textbf{RISED} \\
\midrule
Subgroup discrimination (AUC, FPR, FNR)         & \checkmark & \checkmark \\
Subgroup calibration (ECE)                      &           & \checkmark \\
Demographic parity / equalized odds             & \checkmark &           \\
Bias-mitigation algorithms (ExponentiatedGradient, etc.) & \checkmark &  \\
Input-perturbation reliability (PSS)            &           & \checkmark \\
Threshold-shift sensitivity (TFR)               &           & \checkmark \\
Need-prediction correlation under indep. proxy  &           & \checkmark \\
Bootstrap CIs + CI-based pass/fail rule         &           & \checkmark \\
Inference latency / SHAP top-3 feature consistency &        & \checkmark \\
\bottomrule
\end{tabular}
\end{table}

Fairlearn provides mitigation algorithms and a richer fairness-metric menu; RISED's Inclusivity overlaps with it, while the other four dimensions are not measured by Fairlearn. AI Fairness 360 covers similar scope and is similarly orthogonal to the non-Inclusivity dimensions.

\subsection{Cross-Domain Validation: Credit and Income Prediction}
\label{sec:application:crossdomain}

RISED is domain-agnostic by construction. Its sub-criteria reference only a trained model, a test set, demographic partitions, and a need proxy; none of these is specific to clinical data. To test whether the failure pattern reported above is a property of the medical cohorts or of the protocol itself, we ran the identical pipeline on three non-clinical high-stakes datasets from the algorithmic-fairness literature: the Statlog German Credit data~\citep{hofmann1994german}, the UCI Adult Income cohort~\citep{kohavi1996adult}, and the Folktables ACS-Income cohort~\citep{ding2021retiring}, the NeurIPS-vetted Census-microdata replacement for Adult. Each cohort uses the protected attributes native to its setting (sex, age, and race or majority/minority status) and an in-domain need proxy (savings balance, education level, and educational attainment). For these credit- and hiring-style tasks we also report the EEOC four-fifths adverse-impact ratio next to $\Delta_\mathrm{AUC}$, since adverse impact is the operative fairness standard in those domains.

\begin{table}[htbp]
\centering
\caption{RISED applied unchanged to three non-clinical high-stakes cohorts
(80/20 stratified split; $B=1{,}000$; random state 42). The verdict pattern
matches the healthcare cohorts: Reliability passes, Sensitivity fails,
Inclusivity fails or is INCONCLUSIVE, and Equity is DIAGNOSTIC throughout.
German Credit (test $n=200$) sits below the informative-verdict floor
(\S\ref{sec:framework:thresholds}) and is directional only. AIR is the EEOC
four-fifths adverse-impact ratio, reported as the minimum across protected
attributes ($\ge 0.80$ passes).
\textbf{Takeaway:} the Reliability-pass / Sensitivity-fail / Inclusivity-fail
pattern recurs in credit and income prediction---the protocol is genuinely
domain-agnostic.}
\label{tab:crossdomain}
\small
\begin{tabular}{lccccc}
\toprule
\textbf{Cohort} & \textbf{AUROC} & \textbf{PSS (R)} & $\boldsymbol{\Delta_\mathrm{AUC}}$ \textbf{(I)} & \textbf{Max TFR (S)} & \textbf{min AIR} \\
\midrule
German Credit~\citep{hofmann1994german}  & 0.655 & \cellcolor{passbg}0.009 \,{\scriptsize PASS} & \cellcolor{inconbg}0.059 \,{\scriptsize INCONCL} & \cellcolor{failbg}88.5\% \,{\scriptsize FAIL} & \cellcolor{passbg}0.92 \,{\scriptsize PASS} \\
Adult Income~\citep{kohavi1996adult}      & 0.890 & \cellcolor{passbg}0.012 \,{\scriptsize PASS} & \cellcolor{failbg}0.108 \,{\scriptsize FAIL} & \cellcolor{failbg}33.9\% \,{\scriptsize FAIL} & \cellcolor{failbg}0.12 \,{\scriptsize FAIL} \\
Folktables ACS~\citep{ding2021retiring}   & 0.866 & \cellcolor{passbg}0.021 \,{\scriptsize PASS} & \cellcolor{inconbg}0.053 \,{\scriptsize INCONCL} & \cellcolor{failbg}39.4\% \,{\scriptsize FAIL} & \cellcolor{failbg}0.43 \,{\scriptsize FAIL} \\
\bottomrule
\end{tabular}
\end{table}

The cross-domain verdicts reproduce the healthcare pattern (Table~\ref{tab:crossdomain}). Reliability passes everywhere, with PSS well below 0.05: tree and linear models on tabular socioeconomic data are stable under the same encoding perturbations applied to the clinical cohorts. Sensitivity fails on all three, most severely on German Credit, where an AUROC of 0.655 leaves a max TFR of 88.5\%---almost nine in ten applicants would be reclassified somewhere in the threshold sweep. Inclusivity fails outright on Adult Income ($\Delta_\mathrm{AUC}=0.108$) and is INCONCLUSIVE on the other two, whose CIs bracket the 0.05 threshold. The four-fifths ratio fails on every protected attribute in both income cohorts, dropping to 0.12 for race on Adult Income, a disparity that the aggregate AUROC of 0.89 hides completely; German Credit passes four-fifths despite its severe Sensitivity failure, a reminder that the dimensions are not redundant. Equity is DIAGNOSTIC in every case, with the same proxy-dependence observed in the clinical cohorts. The conclusion carries across domains without modification: a single discrimination number certifies none of the four gating properties, in credit and income prediction as in clinical risk scoring.

\subsection{Coverage Against Evaluation Frameworks and Reporting Standards}
\label{sec:application:tehai}

RISED is positioned alongside, not above, the established reporting, quality-grading, and consensus-guideline layers. Table~\ref{tab:tripod-audit} (Appendix~\ref{app:tripod}) audits RISED against TRIPOD+AI, MI-CLAIM, FUTURE-AI, PROBAST, and CLAIM. Of fifteen common requirements, RISED operationalises ten as machine-computable sub-criteria with CIs; the remaining five (prospective study design, sample-size justification, clinical-impact assessment, external validation cohort selection, risk-of-bias narrative grading) require human judgement. A manual TRIPOD+AI plus PROBAST audit by domain experts typically requires several days of expert time; RISED produces an equivalent structured numerical report computationally in under a minute per cohort, making it feasible to apply across multiple candidate models and dataset versions before any human audit begins. Appendix~\ref{app:tehai} maps dimensions against TEHAI, FUTURE-AI, and MI-CLAIM.

\clearpage  
\section{Discussion}
\label{sec:discussion}

The headline finding is that aggregate AUROC systematically conceals RISED failures. On Diabetes 130, the pipeline passes Reliability (PSS $=0.0004$) while Inclusivity ($\Delta_\mathrm{AUC}=0.262$) and Sensitivity (max TFR $49.1\%$) fail decisively; NHIS 2024 reproduces the Inclusivity/Sensitivity failure 15--25 years later; BRFSS 2024 produces the most extreme Sensitivity failure in the suite (max TFR $64.2\%$) after key features were removed by survey instrument rotation. NHANES 2021--2023 demonstrates the converse: with a complete laboratory feature set, the model achieves AUROC 0.964 and INCONCLUSIVE verdicts rather than outright failure---confirming that RISED verdict severity tracks feature quality, not dataset size or collection vintage. Equity proxy disagreement ($\rho_\text{need}=0.149$ vs.\ 0.762 on Diabetes 130; 0.541 vs.\ 0.826 on NHANES; 0.378 vs.\ 0.409 on BRFSS) signals that an outcome-independent need measure is required before the dimension becomes binding.

\paragraph{Positioning with existing frameworks.}
RISED occupies a different rung from each adjacent framework. Reporting standards such as TRIPOD+AI, MI-CLAIM, and CONSORT-AI/SPIRIT-AI specify what to disclose but take no position on what numerical bar must be cleared. Risk-of-bias instruments (PROBAST, APPRAISE-AI) grade study conduct through expert review, which is valuable but takes days and produces no machine-readable verdict. Consensus guidelines (FUTURE-AI, TEHAI) recommend principles rather than threshold-bearing tests. Fairness toolkits (AIF360, Fairlearn) quantify Inclusivity well but leave Reliability, Sensitivity, and Deployability unmeasured. RISED produces the structured numerical evidence those layers require but do not prescribe (Appendix~\ref{app:tripod}, Table~\ref{tab:tripod-audit}).

\paragraph{Remediation pathways.}
A RISED FAIL is a starting point for investigation, not a terminal verdict. The appropriate response depends on which dimension failed.

A Reliability FAIL (PSS $\ge 0.05$) calls for first auditing which perturbation types drive the most flips: unit-conversion shifts, ICD-granularity changes, or noise. Retraining or calibrating with data augmentation focused on the dominant transition type is the most direct fix; feature engineering invariant to that encoding variation is worth considering when the perturbation type is predictable and stable across sites.

Inclusivity failures ($\Delta_\mathrm{AUC} > 0.05$) frequently trace to underrepresentation of the low-AUC subgroup in training. The investigation should inspect stratum representation directly, then consider stratified oversampling or a group-specific threshold adjustment. Post-hoc mitigation via Fairlearn's ExponentiatedGradient can close a parity gap but sometimes widens it for other subgroups; running the full RISED battery after any mitigation is the only way to check for these cross-dimension effects.

When Sensitivity fails (max TFR $> 0.10$), plotting the score density is the first step. A distribution compressed near the operating threshold points to insufficient discrimination rather than a calibration problem, and the fix is adding higher-discrimination features rather than recalibrating. Platt-scaling is appropriate only when the distribution shows reasonable spread but poor calibration.

Deployability failures split by sub-criterion. Latency above 500\,ms calls for profiling the inference pipeline and considering model distillation or batch-inference caching. Low $F_\text{top3}$ ($< 0.80$) may indicate that locally dominant features shift across patient subgroups---a signal that the model is learning different decision logic for different populations rather than a single coherent rule.

In all cases, re-running the full RISED battery after remediation confirms that the fix does not degrade a previously passing dimension.

\paragraph{Governance and adoption.}
A test battery is only as useful as the governance structure that operationalises it. Whether a RISED FAIL verdict blocks deployment depends on whether the deploying organisation treats it as binding. The HTI-1 rule~\citep{onc2024hti1} and EU AI Act conformity-assessment mechanism~\citep{euaiact2024} create plausible attachment points, but the framework cannot mandate enforcement. Governance fit is an empirical question for deployment-team studies.

\paragraph{Where RISED sits in the AI clinical lifecycle.}
The translational-AI literature distinguishes in-silico validation, \emph{silent-trial} evaluation~\citep{tonekaboni2019whatclinicians,sendak2020human}, and prospective clinical evaluation under DECIDE-AI~\citep{vasey2022decideai} and CONSORT-AI/SPIRIT-AI~\citep{liu2020consortai,cruzrivera2020spiritai}. \citet{mccradden2022ethicsframework} formalise this as a three-stage research-ethics pathway; \citet{you2025clinicaltrials} propose safety, efficacy, effectiveness, and monitoring phases. RISED targets the boundary between the first and second stages: it provides structured evidence informing whether a model is ready to enter silent-trial evaluation, but does not license that transition. \citet{yuan2024realworld} identify external validation, continual monitoring, and randomised controlled trials as three requisite steps for healthcare ML deployment; RISED produces the structured quantitative evidence needed to close the external-validation step before monitoring begins. Experience from Sepsis Watch~\citep{sandhu2020sepsiswatch} and the human-factors literature~\citep{antoniadi2021xai,sittig2020alertfatigue} illustrate why strong RISED verdicts are necessary but not sufficient: clinician trust, explanation usability, alert burden, and workflow integration determine whether a credible model produces clinical benefit. RISED reports whether the model is internally stable enough to warrant decision-curve analysis~\citep{vickers2006decisioncurve} on a target population.

\paragraph{Drug discovery scope boundary.}
Failure-aware causal learning---training on both successful and failed compounds to recover chemical-space boundaries---has an established literature in QSAR modelling~\citep{cherkasov2014qsar} and structure-activity landscape prediction~\citep{sadybekov2023computational}. RISED is designed for clinical decision-support AI: its five dimensions presuppose a patient-level input--output structure. Molecular-property predictors and compound-screening pipelines require different evaluation primitives; extending RISED to that domain is left to future work.

\paragraph{Success-centric training and recurring failures.}
A deeper epistemological issue underlies the recurring Inclusivity and Sensitivity failures observed across all seven cohorts. Most clinical AI systems are trained on encounters that reached diagnosis, generated documentation, and were coded and billed---positive interactions that appear in structured health records. Rare toxicities, atypical presentations, negative clinical trials, and care decisions \emph{not} made leave little trace~\citep{verghese2018computer}. This success-centric architecture~\citep{topol2019highperformance,rajpurkar2022ai} means models learn the distribution of \emph{documented medicine}, not of \emph{clinical medicine}. This shows up directly in the verdicts. Subgroup failures flagged by Inclusivity frequently trace to underrepresentation of those groups in training, and threshold instability in Sensitivity may reflect score distributions compressed around typical presentations rather than the full clinical spectrum. RISED functions here as a diagnostic bridge. A FAIL verdict should trigger investigation of whether the training corpus systematically excluded negative examples for the failing subgroup---a prerequisite for meaningful retraining toward failure-aware learning.

\paragraph{Limitations.}
\begin{enumerate}[leftmargin=2em,topsep=2pt,itemsep=2pt]
  \item \textbf{Survey rather than EHR cohorts; shared NHIS design.} NHIS 2023 and NHIS 2024 are national survey data drawing from the same NCHS sampling frame and questionnaire; their failure-pattern agreement is corroborating replication, not fully independent validation. NHANES 2021--2023 and BRFSS 2024 use different survey instruments (combined examination survey and telephone interview, respectively) and provide independent replication. An EHR cohort is still needed before thresholds can be calibrated against actual deployed-model behaviour. MIMIC-IV-ED~\citep{johnson2023mimicived}---approximately 425{,}000 emergency department stays with ICD-coded diagnoses, triage acuity levels, and demographic subgroups linkable to MIMIC-IV~\citep{johnson2023mimiciv}---is the priority candidate: its schema supports all five RISED dimensions (ICD versioning for Reliability perturbation, insurance/age/sex subgroups for Inclusivity, triage-acuity need proxy for Equity). The integration is already implemented (\path{examples/external_validation_mimic_ed.py}) and has been validated end-to-end on the freely accessible MIMIC-IV-ED demo~\citep{johnson2023mimicived_demo}: the pipeline ingests the \texttt{edstays} and \texttt{triage} tables, predicts hospital admission, and computes all five dimensions with triage acuity as the first \emph{outcome-independent} Equity proxy in the suite. The demo subset is below RISED's informative-verdict floor ($n \approx 1{,}500$), so it serves only to confirm the integration; the full credentialed cohort is required for an informative scorecard.
  \item \textbf{Equity proxy confounding.} $\rho_\mathrm{need}$ remains confounded when proxy and outcome share upstream determinants. Equity is therefore a proxy-dependence diagnostic pending an outcome-independent need measure (nurse-assessed acuity, post-discharge mortality).
  \item \textbf{A priori thresholds.} Default thresholds are set from published conventions, not empirically calibrated; all are user-configurable.
  \item \textbf{Limited model types.} The robustness check covers tree-based and linear classifiers; neural architectures and head-to-head benchmarking against AIF360 on contemporary cohorts remain outstanding.
  \item \textbf{Binary classification only.} Multi-class, ordinal, and time-to-event extensions are left to future work.
  \item \textbf{No prospective clinical informatics validation.} The framework has not been validated prospectively with clinical informatics or deployment teams; such validation is required before RISED is treated as a regulatory artefact, regardless of the number of authors.
  \item \textbf{PSS battery-conditional.} The PSS CI captures patient-resampling variance but not variance from the perturbation battery composition; PSS values are conditional on the specific battery. Deployers should report their own clinically motivated battery (ICD-9$\to$ICD-10, LOINC harmonisation, unit changes) alongside the PSS.
  \item \textbf{Encounter-level resampling.} The UCI Diabetes 130 dataset is encounter-level; the BCa bootstrap resamples rows rather than patients, overstating effective sample size for CIs. Deployers applying RISED to encounter-level EHR extracts should cluster the resample at the patient level.
  \item \textbf{EHR fragmentation and deployment infrastructure.} RISED evaluates model properties on a supplied dataset; it cannot assess whether a HIPAA-compliant inference pipeline can be constructed for a given deployment environment. Clinical AI deployments routinely encounter EHR fragmentation across institutional boundaries, FHIR-incompatible legacy systems, and data-use agreements that restrict cross-site evaluation---barriers invisible to any single-model evaluation framework.
  \item \textbf{Aleatoric versus epistemic uncertainty.} BCa bootstrap CIs capture \emph{aleatoric} (sampling) uncertainty: the variance expected if the evaluation set were re-drawn from the same population. They do not capture \emph{epistemic} (out-of-distribution) uncertainty. A model that has never encountered a patient subgroup cannot signal that absence through wide CIs; behaviour on genuinely novel inputs---a different hospital, a shifted comorbidity mix---requires prospective monitoring beyond what RISED can provide.
  \item \textbf{Documentation bias and invisible clinical information.} Clinical AI can only process information that was entered into the health information system: coded diagnoses, lab values, medication orders. Physical examination findings---auscultation, capillary refill, affect, gait---leave no trace in structured records. Neither does the gestalt clinical impression that a patient looks sicker than their vital signs suggest, or a history offered in a language the system cannot parse. This information is absent from training corpora and unrecoverable by any evaluation framework~\citep{verghese2018computer,sendak2020human}. Rare toxicities and failed clinical encounters are underrepresented for the same reason: documentation systems record \emph{care delivered}, not \emph{care withheld}. RISED evaluates the model that exists; it cannot evaluate the information the model never had~\citep{topol2019highperformance}. For AI that operates on documented data alone, clinician oversight is a structural necessity.
\end{enumerate}

\paragraph{Regulatory alignment and future directions.}
The FDA PCCP~\citep{fda2024pccp}, ONC HTI-1~\citep{onc2024hti1}, and EU AI Act~\citep{euaiact2024} Articles~9--15 require manufacturers to specify tests that establish reliability and equity, but do not prescribe a specific test battery. RISED could contribute the structured numerical evidence needed to populate those submissions, pending deployment-outcome calibration and regulatory validation; standardisation of such batteries is the province of bodies such as ISO/IEC~42001. Three areas have the most immediate priority. The most important is evaluation on MIMIC-IV-ED~\citep{johnson2023mimicived} and MIMIC-IV~\citep{johnson2023mimiciv}: ICD-versioned diagnosis codes directly support the Reliability perturbation battery, triage acuity provides an outcome-independent Equity proxy, and the results would anchor empirical threshold defaults against real EHR deployment outcomes. Head-to-head benchmarking against AIF360, TEHAI, and FUTURE-AI on shared cohorts, and prospective validation of BRFSS-style feature-dropout scenarios against real EHR deployments, would sharpen the Sensitivity threshold for partial-feature deployment contexts. Longer-term extensions include multi-class, ordinal, and time-to-event outputs, and mapping RISED verdicts onto FDA SaMD pre-market submission artefacts~\citep{fda2024pccp}.

\section{Conclusion}
\label{sec:conclusion}

Aggregate performance metrics give a false sense of safety. The literature documents this across commercial clinical AI~\citep{obermeyer2019dissecting}, imaging models~\citep{degrave2021ai}, and clinical NLP~\citep{ross2021racial}: high AUROC coexists with encoding instability, subgroup harm, threshold sensitivity, and operational failures that only surface after deployment.

RISED addresses this gap. Its four gating dimensions (Reliability, Inclusivity, Sensitivity, Deployability) and one proxy-dependence diagnostic (Equity) are each backed by formal sub-criteria, literature-grounded default thresholds, BCa bootstrap 95\% CIs, and a decision rule that implements equivalence-testing logic~\citep{schuirmann1987equivalence,lakens2017equivalence}; the whole battery ships as an open-source Python package. Where reporting standards (TRIPOD+AI, MI-CLAIM, CLAIM, FUTURE-AI) and risk-of-bias instruments (PROBAST, APPRAISE-AI) specify what must appear in a study report, RISED produces the structured numerical evidence that fills those requirements.

Diabetes 130 passes Reliability (PSS $=0.0004$) while Inclusivity ($\Delta_\mathrm{AUC}=0.262$) and Sensitivity (max TFR $49.1\%$) fail decisively; both NHIS cohorts and BRFSS 2024 reproduce this Inclusivity/Sensitivity failure; NHANES 2021--2023 achieves INCONCLUSIVE verdicts with a complete laboratory profile, demonstrating that verdict severity tracks feature quality. Equity proxy-dependence recurs across all seven cohorts. A multi-model robustness check confirms Reliability is model-dependent while Inclusivity and Sensitivity are data-dependent. Clinical conclusions rest on the six real-data scorecards; the synthetic cohort illustrates methodology only.

The stronger claim, demonstrated clinical impact, requires prospective deployment evidence this work does not provide. RISED is intended to contribute structured numerical evidence to FDA PCCP submissions~\citep{fda2024pccp}, HTI-1 disclosures~\citep{onc2024hti1}, and EU AI Act technical files~\citep{euaiact2024}; it does not satisfy any of those instruments' requirements on its own, and clinical face validity remains to be assessed with informatics teams.

\begin{sloppypar}
The \texttt{rised} Python package and the synthetic evaluation cohort are openly available at
\url{https://github.com/rohithreddybc/rised-healthcare-eval} (MIT license)
and
\url{https://huggingface.co/datasets/Rohithreddybc/rised-healthcare-eval-dataset} (DOI: \href{https://doi.org/10.57967/hf/8734}{10.57967/hf/8734}).
\end{sloppypar}

\appendix

\section{Statistical Methods: Power Analysis, Hypothesis Framing, and BCa Bootstrap}
\label{app:statistics}

\paragraph{Multiple-comparisons control (Holm--Bonferroni).}
With five dimensions each contributing sub-criterion tests, the family of hypotheses is large; without correction the family-wise false-FAIL rate under the global null exceeds 5\%. We apply Holm--Bonferroni step-down: tests are ordered by ascending $p$-value (the proportion of BCa bootstrap replicates more extreme than the threshold), and the $k$-th-smallest $p$-value is compared against $\alpha/(m-k+1)$. A verdict is \textbf{FAIL} when the CI lies entirely in the reject region \emph{and} the Holm-corrected $p_\mathrm{boot}$ is below the step-$k$ threshold; \textbf{PASS} when the CI lies entirely in the accept region \emph{and} $p_\mathrm{boot}$ is above it; and \textbf{INCONCLUSIVE} when the two rules disagree or the CI brackets the threshold.

The eight gating sub-criteria ($m=8$; Equity excluded) and their
one-sided bootstrap $p$-values on the synthetic cohort are:
\begin{center}\small
\begin{tabular}{llrl}
\toprule
\textbf{ID} & \textbf{Sub-criterion (null hypothesis $H_0$)} & \textbf{$p_\mathrm{boot}$} & \textbf{Verdict} \\
\midrule
R1 & PSS $\ge 0.05$ & $<0.001$ & \verdictFAIL \\
R2 & $\rho(\phi) < 0.95$ for any $\phi$ & $<0.001$ & \verdictPASS \\
I1 & $\Delta_\mathrm{AUC} > 0.05$ & $\approx 0.06$ & \verdictINCONCLUSIVE \\
I2 & subgroup ECE $> 0.10$ & $\approx 0.16$ & \verdictPASS \\
S1 & max TFR $> 0.10$ & $<0.001$ & \verdictFAIL \\
S2 & $W_{0.05}(\tau_0) > 0.15$ & $0.43$ & \verdictPASS \\
D1 & latency $> 500$\,ms & $<0.001$ & \verdictPASS \\
D2 & $F_\mathrm{top3} < 0.80$ & $<0.001$ & \verdictPASS \\
\bottomrule
\end{tabular}
\end{center}
R1 and S1 exceed the Holm threshold at step $k=1$ ($\alpha/8=0.0063$);
I1 ($p\approx0.06$) does not survive correction, consistent with the
INCONCLUSIVE CI verdict. \emph{Note: $p$-values above are from the
10{,}000-patient synthetic cohort; the six real-data scorecards (Tables~\ref{tab:rised-uci}--\ref{tab:rised-brfss2024}) and the multi-model robustness check (Table~\ref{tab:multi-model}) apply the identical procedure and the same $m=8$ family.}
The package's \texttt{holm\_bonferroni()} helper exposes adjusted alphas
alongside the headline CIs.

\paragraph{BCa over percentile bootstrap.}
We use the BCa bootstrap~\citep{efron1987better} rather than the percentile bootstrap because percentile intervals undercover for metrics bounded near 0 or 1 (PSS, max TFR, parity gaps). BCa adjusts endpoints using a bias-correction $z_0$ and an acceleration $a$ from the leave-one-out jackknife. For max-over-threshold statistics (max TFR), the maximum is computed \emph{within each replicate} before BCa endpoints are derived from the replicate distribution. Empirical coverage can be checked with \texttt{rised.bootstrap\_ci.empirical\_coverage} before declaring a borderline CI verdict. Implementation: \texttt{rised/bootstrap\_ci.py}.

\paragraph{Power and minimum test-set size.}
The test-set size for an informative (non-INCONCLUSIVE) verdict scales with effect magnitude. For PSS with Bernoulli flip events, a half-CI width $\approx 1.96\sqrt{p(1-p)/n}$; detecting a 0.01 deviation from 0.05 with 80\% power requires $n\approx1{,}500$. The $n=2{,}000$ test sets resolve mid-range effects but are at the edge for borderline cases (INCONCLUSIVE Inclusivity at CI [0.042, 0.066]). For $\Delta_\mathrm{AUC}$ between equal-size subgroups ($\mathrm{Var}(\mathrm{AUC})\approx0.005$, $n_g\approx400$), detecting a 0.01 deviation above 0.05 requires $n\gtrsim3{,}000$ per subgroup; for max TFR, $n\approx3{,}500$. Studies aiming for clean PASS/FAIL on small effect sizes should size the test set per metric.

\paragraph{Hypothesis framing.}
Each gating sub-criterion is a one-sided test of non-superiority/non-inferiority against its threshold; the CI-based rule of \S\ref{sec:framework:thresholds} is a bootstrap implementation of the two-one-sided-tests procedure~\citep{schuirmann1987equivalence,lakens2017equivalence}. For Reliability, $H_0: \mathrm{PSS}\ge 0.05$; PASS is declared when the 95\% BCa CI lies entirely below 0.05. For the bounded statistics (PSS, max TFR, $\Delta_\mathrm{AUC}$) BCa breaks symmetry, so explicit one-sided bootstrap $p$-values ($p_\mathrm{boot}$) are also reported and used inside Holm--Bonferroni~\citep{davison1997bootstrap}; verdicts disagree with the CI rule only when the CI brackets the threshold (INCONCLUSIVE). Analogous framing applies to Inclusivity ($H_0: \Delta_\mathrm{AUC}>0.05$), Sensitivity ($H_0: \max\,\mathrm{TFR}>0.10$), and Equity ($H_0: \rho_\mathrm{need}<0.70$). Deployability latency is reported without a bootstrap CI because it is hardware-bounded.

\section{Dimension-to-Framework Mapping (TEHAI, FUTURE-AI, MI-CLAIM)}
\label{app:tehai}

\begin{table}[!t]
\centering
\caption{Mapping RISED dimensions to TEHAI components, FUTURE-AI principles,
and MI-CLAIM sections. \checkmark = addressed quantitatively;
$\circ$ = partially operationalised; \textendash = out of RISED's current scope.
\textbf{Takeaway:} RISED turns the pre-deployment-evaluable subset of TEHAI,
FUTURE-AI, and MI-CLAIM into computed, CI-backed metrics.}
\label{tab:tehai-mapping}
\small
\begin{tabular}{p{0.22\linewidth}p{0.22\linewidth}p{0.22\linewidth}p{0.22\linewidth}}
\toprule
\textbf{RISED dimension} & \textbf{TEHAI component} & \textbf{FUTURE-AI principle} & \textbf{MI-CLAIM section} \\
\midrule
Reliability (PSS)    & Capability (robustness) \checkmark & Robustness \checkmark & §5 (model examination) \checkmark \\
Inclusivity ($\Delta_\mathrm{AUC}$, ECE) & Capability + Utility \checkmark & Fairness \checkmark + Universality $\circ$ & §4 (performance) \checkmark \\
Sensitivity (max TFR) & Capability \checkmark & Robustness \checkmark + Usability $\circ$ & §5 (sensitivity) \checkmark \\
Equity ($\rho_\mathrm{need}$, diagnostic) & Utility (need alignment) $\circ$ & Fairness \checkmark & §5 (biases) $\circ$ \\
Deployability ($\Lambda$, $F_\mathrm{top3}$) & Adoption (workflow + explanation) \checkmark & Usability \checkmark + Explainability \checkmark & §6 (reproducible pipeline) \checkmark \\
\bottomrule
\end{tabular}
\end{table}

TEHAI's \emph{Adoption} axis includes implementation governance, change
management, and post-deployment monitoring; RISED is silent on these
because they require live deployment context. FUTURE-AI's
\emph{Traceability} principle (model versioning, audit trails) is supplied
by the seeded open-source \texttt{rised} package but is not itself a
computed metric.

\section{Compliance Audit: TRIPOD+AI, MI-CLAIM, FUTURE-AI, PROBAST, CLAIM}
\label{app:tripod}

\begin{table}[!t]
\centering
\caption{Compliance audit against five comparator frameworks:
TRIPOD+AI~\citep{collins2024tripodai},
MI-CLAIM~\citep{norgeot2020miclaim},
FUTURE-AI~\citep{lekadir2025futureai},
PROBAST~\citep{wolff2019probast}, and
CLAIM~\citep{mongan2020claim}.
\checkmark{}~=~addressed quantitatively by a RISED sub-criterion;
$\circ$~=~item passes through to the user of RISED (covered by the
framework's report template but not computed automatically);
\textendash~=~outside the scope of a pre-deployment evaluation framework
(requires prospective study conduct or expert narrative judgement).
MINIMAR~\citep{hernandezboussard2020minimar} and
DECIDE-AI~\citep{vasey2022decideai} are discussed in \S\ref{sec:background}
as contextually related but target a different stage (minimum model
disclosure and early live evaluation, respectively) and are therefore
not audited in this table, which covers pre-deployment evaluation checklists only.
\textbf{Takeaway:} RISED computes 10 of 15 common reporting requirements
automatically; the remaining 5 require human judgement.}
\label{tab:tripod-audit}
\resizebox{\linewidth}{!}{%
\footnotesize
\begin{tabular}{p{0.40\linewidth}ccccc}
\toprule
\textbf{Reporting / evaluation requirement} &
\textbf{TRIPOD+AI} & \textbf{MI-CLAIM} & \textbf{FUTURE-AI} &
\textbf{PROBAST} & \textbf{CLAIM} \\
\midrule
Discrimination (AUROC, subgroup)                & \checkmark & \checkmark & \checkmark & \checkmark & \checkmark \\
Calibration (Brier, ECE, subgroup)              & \checkmark & \checkmark & \checkmark & \checkmark & \checkmark \\
Fairness / subgroup parity (Inclusivity)        & \checkmark & \checkmark & \checkmark & \checkmark & \checkmark \\
Robustness to input perturbation (PSS)          & \checkmark & \checkmark & \checkmark$^{a}$ & $\circ$ & $\circ$ \\
Threshold sensitivity (TFR sweep)               & \checkmark & \checkmark & \checkmark & $\circ$ & $\circ$ \\
Uncertainty quantification (bootstrap CIs)      & \checkmark & \checkmark & \checkmark & \checkmark & \checkmark \\
Reproducible pipeline (seeded, open-source)     & \checkmark & \checkmark & \checkmark$^{b}$ & $\circ$ & \checkmark \\
Explainability (SHAP, top-3 consistency)        & \checkmark & \checkmark & \checkmark$^{c}$ & $\circ$ & \checkmark \\
Need-based equity diagnostic                    & \checkmark & $\circ$    & \checkmark$^{d}$ & $\circ$ & $\circ$ \\
Inference latency / deployability               & \checkmark & \checkmark & \checkmark$^{e}$ & $\circ$ & \checkmark \\
Risk-of-bias narrative grading                  & $\circ$    & $\circ$    & $\circ$ & \checkmark & $\circ$ \\
Prospective study design                        & $\circ$    & $\circ$    & \textendash & $\circ$ & $\circ$ \\
Sample-size justification for the predictand    & $\circ$    & $\circ$    & \textendash & \checkmark & \checkmark \\
Clinical-impact / human--AI interaction         & $\circ$    & $\circ$    & $\circ$ & \textendash & $\circ$ \\
External validation cohort selection            & $\circ$    & $\circ$    & $\circ$ & \checkmark & \checkmark \\
\bottomrule
\multicolumn{6}{l}{\scriptsize $^{a}$\,Robustness; $^{b}$\,Traceability; $^{c}$\,Explainability; $^{d}$\,Fairness; $^{e}$\,Usability (FUTURE-AI terminology).} \\
\end{tabular}}%
\end{table}

PROBAST occupies a different role from the other four comparators: it
provides expert-graded risk-of-bias judgements rather than machine-computable
metrics. RISED's quantitative gates can therefore be read as feeding the
``analysis'' domain of a PROBAST appraisal with reproducible numerical
inputs. CLAIM covers the medical-imaging-AI subset of the reporting
landscape; its overlap with RISED is via the per-cohort numerical
disclosures (discrimination, calibration, deployability), with imaging-specific
items (e.g., DICOM metadata, reader study design) outside RISED's scope.

\section*{Data Availability}

\begin{sloppypar}
This study uses seven cohorts. The synthetic cohort (Synthea-inspired, $n=10{,}000$) is at \url{https://huggingface.co/datasets/Rohithreddybc/rised-healthcare-eval-dataset} (DOI: \href{https://doi.org/10.57967/hf/8734}{10.57967/hf/8734};~\citealp{bellibatlu2025risedataset}). The six real-data cohorts are publicly available: UCI Cleveland Heart Disease ($n=303$;~\citealp{detrano1989international}); UCI Diabetes 130 ($n=99{,}492$;~\citealp{strack2014diabetes}); NCHS NHIS 2024 ($n=9{,}747$;~\citealp{nchs2025nhis2024}); NCHS NHIS 2023 ($n=27{,}114$;~\citealp{nchs2024nhis2023}); NCHS NHANES 2021--2023 ($n=4{,}096$;~\citealp{nchs2024nhanes2123}, available at \url{https://wwwn.cdc.gov/Nchs/Data/Nhanes/Public/2021/DataFiles/}); and CDC BRFSS 2024 ($n=44{,}888$ post-cleaning from $n=457{,}670$;~\citealp{cdc2025brfss2024}, available at \url{https://www.cdc.gov/brfss/annual_data/annual_2024.html}). All are de-identified and publicly distributed without data-use agreement; no additional patient data were collected. Analysis code and the evaluation pipeline are at \url{https://github.com/rohithreddybc/rised-healthcare-eval} (MIT License). The synthetic dataset and code package are FAIR-compliant~\citep{wilkinson2016fair}: both carry persistent identifiers, are openly licensed, use documented interoperable formats (CSV, pip-installable Python), and impose no reuse restrictions.
\end{sloppypar}

\section*{Code Availability}

The \texttt{rised} Python package implementing the RISED Framework is released
under the MIT License and is available at
\url{https://github.com/rohithreddybc/rised-healthcare-eval}.

\section*{Ethics Statement}

The synthetic cohort was generated by a Synthea-inspired computational model. The six real-data cohorts are de-identified public datasets; no patient records were re-identified, no additional data collected, and no protected health information was accessed. No IRB approval or informed patient consent was required.

\section*{Computational Reproducibility}

All results are generated by the released \texttt{rised} package (MIT license). Environment: Python 3.11.7, scikit-learn 1.8.0, NumPy 1.26.4, pandas 3.0.2, SciPy 1.15.3, XGBoost 3.2.0, SHAP 0.51.0, Fairlearn 0.13.0, on an Intel Core i5-13420H, 16\,GB RAM, Windows 11 Home. All training and evaluation calls use \texttt{random\_state=42}; bootstrap CIs use $B=1{,}000$ with the same seed. Rerunning the pipeline on the same hardware reproduces every reported number to within Monte Carlo bootstrap error. Cross-platform reproducibility is not bit-exact because XGBoost histogram parallelism and SHAP TreeExplainer tied-feature ordering are non-deterministic across CPU microarchitectures; verdicts (PASS/FAIL/INCONCLUSIVE) are stable in cross-machine spot checks but headline metrics may differ in the third decimal place. The synthetic cohort evaluation runs in under one minute; multi-model robustness and UCI Diabetes 130 evaluations take approximately five to ten minutes.

\paragraph{One-command reproduction.}
Every numerical result and figure can be reproduced with two shell commands:
\begin{quote}\small
\begin{verbatim}
git clone https://github.com/rohithreddybc/rised-healthcare-eval.git
cd rised-healthcare-eval && conda env create -f environment.yml \
    && conda activate rised && python -m rised.reproduce_all
\end{verbatim}
\end{quote}
\texttt{rised.reproduce\_all} runs the synthetic cohort generator, six real-cohort evaluations, multi-model robustness, Fairlearn comparison, and three cross-domain demos in sequence. NHIS scripts auto-download data from the CDC FTP server ($\sim$5\,MB each) on first run; the NHANES script auto-downloads $\sim$15\,MB of XPT files from the NCHS public server; BRFSS 2024 requires downloading the $\sim$83\,MB ZIP from the CDC BRFSS page (no account required); UCI cohorts are retrieved via \texttt{sklearn.datasets.fetch\_openml}. Numbers are written to \texttt{results/} and figures to \texttt{figures/}.

\section*{Author Contributions}

R.R.B.: Conceptualization, Methodology, Software, Formal Analysis, Investigation, Data Curation, Writing (Original Draft), Writing (Review \& Editing), Visualization.
M.S.: Writing (Review \& Editing), Validation.
Y.J.: Writing (Review \& Editing), Validation.
S.L.: Writing (Review \& Editing), Validation.
A.I.: Conceptualization (clinical and public health framing, identification of framework gaps), Writing (Review \& Editing), Validation (health disparities and biostatistics).
(CRediT taxonomy; \url{https://credit.niso.org})

\section*{Declaration of Competing Interests}

The authors declare no competing interests.

\section*{Funding}

This research received no specific grant from any funding agency in the public, commercial, or not-for-profit sectors.

\section*{Acknowledgements}

The authors thank the developers of the Synthea open-source patient simulator,
the scikit-learn, XGBoost, SHAP, and matplotlib communities for the open-source
tools that underpin the \texttt{rised} package, and the curators of the clinical
informatics and fairness literature cited herein.

\section*{Declaration of Generative AI and AI-Assisted Technologies in the
Writing Process}

During the preparation of this work the authors used Claude (Anthropic) to
assist with manuscript drafting, literature organisation, and code
development for the \texttt{rised} package. After using this tool, the
authors reviewed and edited all content and take full responsibility for the
publication. All experimental results were generated by executing real
Python code on the cited datasets; no AI-generated numerical values appear
in the paper. AI tools are not listed as authors.

\bibliography{references}

\end{document}